\documentclass[11pt, a4paper]{googlecloud}

\pdftrailerid{}
\usepackage[authoryear, sort&compress, round]{natbib}

\usepackage{url}
\usepackage{graphicx}
\usepackage{booktabs}
\usepackage[most]{tcolorbox}
\usepackage{multirow}
\usepackage{algorithm}
\usepackage{algorithmic}
\usepackage{colortbl}
\usepackage{array}
\usepackage{caption}
\usepackage{amssymb}
\usepackage{enumitem}
\definecolor{GoogleBlue}{HTML}{4285F4}
\definecolor{GoogleRed}{HTML}{EA4335}
\definecolor{GoogleYellow}{HTML}{FBBC05}
\definecolor{GoogleGreen}{HTML}{34A853}

\title{Procedural Graphs: Self-Evolving Execution Structures for LLM Agents}

\correspondingauthor{\footnotesize{Corresponding author: soarik@google.com}\\}

\author[1,2,3]{Yuxing Lu}
\author[1]{Yicheng Chen}
\author[1]{Shanchan Wu}
\author[1]{Sercan \"{O}. Ar{\i}k}

\affil[1]{Google}
\affil[2]{Georgia Institute of Technology}
\affil[3]{Peking University}

\begin{abstract}
Large language models are increasingly deployed as agents that plan over long horizons and act through external tools. Most agents select actions through unconstrained generation over an accumulating history, leaving implicit the procedural knowledge of what to do, in what order, and under which conditions. As trajectories lengthen, agents can lose track of their objectives, invoke tools out of order, and repeat unproductive actions. We introduce the \textbf{Procedural Graph}: just as a knowledge graph organizes factual knowledge into (entity, relation, entity) triplets for \textit{what-is} questions, a Procedural Graph organizes procedural knowledge into (procedure, relation, procedure) triplets for \textit{what-to-do} questions. At each decision step, the framework localizes the agent's active node, and a guidance model translates the surrounding subgraph into step-level situational guidance that biases the solver's next action without dictating it. The graph is self-evolving: an LLM refiner contrasts failed trajectories with successful ones and edits the graph's topology and attributes, committing edits that preserve or improve held-out validation performance while retaining rejected ones to discourage repetition. Starting from a minimal skeleton, the loop builds graphs that match or surpass hand-designed ones. It can also repair a flawed expert prior. Across multiple datasets, task types, and LLMs, the Procedural Graph delivers consistent gains over memory-based baselines, and self-evolution further improves performance without manual engineering.
\end{abstract}

\begin{document}

\newtcolorbox{modebox}[1]{
  colback=GoogleBlue!3,
  colframe=GoogleBlue,
  colbacktitle=GoogleBlue,
  coltitle=white,
  fonttitle=\bfseries,
  title=#1,
  arc=3pt,
  outer arc=3pt,
  boxrule=0.8pt,
  left=8pt, right=8pt, top=6pt, bottom=6pt,
  before skip=8pt, after skip=8pt
}

\newtcolorbox{tracebox}[3][]{
  colback=#2!3,
  colframe=#2,
  colbacktitle=#2,
  coltitle=white,
  fonttitle=\bfseries,
  title=#3,
  title after break={#3 (continued)},
  arc=3pt,
  breakable,
  boxrule=0.8pt,
  outer arc=3pt,
  left=8pt, right=8pt, top=6pt, bottom=6pt,
  before skip=8pt, after skip=8pt,
  #1
}

\maketitle

\section{Introduction}
Large language models (LLMs) are increasingly deployed as autonomous agents that plan over long horizons and act through external tools \citep{sumers2023cognitive,qin2024toolllm}. Most agents make decisions through unconstrained generation conditioned on a flat, growing log of prior actions and observations. This places the burden of procedural coherence on free-form generation: the agent must identify relevant observations, infer which steps remain, and choose an action that respects their dependencies. As trajectories lengthen, agents can lose track of their objectives, invoke tools out of order, and repeat unproductive actions.

Existing approaches provide procedural structure through textual memory, conditional guidelines, and explicit workflows. Memory and self-reflection methods \citep{shinn2023reflexion,zhao2024expel} record past experience as free-form text and retrieve it for reuse in the context. Although these records preserve useful experience, the solver must still reconstruct how it applies to the current step and how it constrains the steps that follow. State-conditioned guidelines \citep{fu2024autoguide} provide more targeted advice, but retrieve rules without explicitly connecting successive procedural steps. Workflows and state machines \citep{xiao2024flowbench,zhang2023don} make those steps explicit and constrain execution, but often require manual design. Automated workflow search reduces manual design effort by optimizing workflow structure offline~\citep{zhang2025aflow}. The remaining challenge is to combine an editable procedure representation with guidance that is conditioned on the agent's current progress.

\begin{figure}[t]
    \centering
    \includegraphics[width=0.95\linewidth]{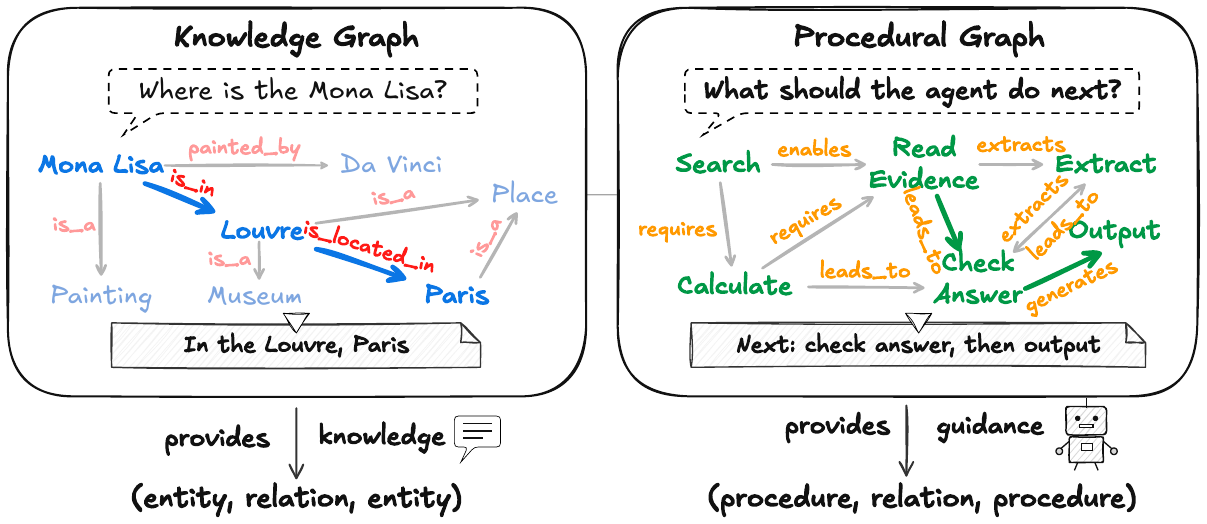}
    \caption{\textbf{From knowledge to procedure.} A Knowledge Graph organizes facts into triplets that answer \textit{what-is} questions. A Procedural Graph organizes task procedures into triplets that answer \textit{what-to-do} questions.}
    \label{fig:kg_vs_pg}
\end{figure}

We argue that an agent needs procedural knowledge that is structured enough to steer it away from invalid behavior, flexible enough to preserve reasoning freedom, responsive to its current progress, and able to improve from experience. We address these requirements with the \textbf{Procedural Graph (PG)}, an explicit and editable directed graph of procedural knowledge. The design mirrors a familiar structure (Figure~\ref{fig:kg_vs_pg}): just as a knowledge graph organizes factual knowledge into (entity, relation, entity) triplets to answer \textit{what-is} questions, a PG organizes procedural information into (procedure, relation, procedure) triplets to answer \textit{what-to-do} questions. Its nodes abstract tool actions, reasoning steps, and states; its edges encode permissible transitions, each annotated with textual attributes describing how and when the transition should be taken. The PG keeps a task domain's procedural knowledge outside the model weights, where it can be inspected, retrieved at each step, and edited without retraining.

Our framework puts this prior to work in two complementary phases. During online inference, the framework localizes the active node from the agent's trajectory, and a guidance model reads the surrounding subgraph in its topological context and translates the relevant edge attributes into situational guidance for the next step. During offline self-evolution, after each batch of training tasks, an LLM refiner contrasts failed trajectories with successful ones and proposes edits to the graph's topology and attributes, adding missing nodes and edges, pruning failure-inducing ones, and revising edge attributes. A structurally valid candidate graph is adopted if it matches or improves performance on a held-out validation set, and rejected candidates are retained as negative constraints. The validation gate filters out candidates that reduce the measured score, while rejection memory discourages repeated unsuccessful proposals.
We summarize our contributions as follows:

\begin{itemize}[leftmargin=*]

\item We introduce the Procedural Graph (PG), an explicit and editable graph of procedural knowledge that steers LLM-agent execution while preserving reasoning flexibility.

\item We propose Generative PG Guidance, an online mechanism that converts the static graph and the live trajectory into step-level situational guidance.

\item We develop a self-evolution loop that refines graph topology and attributes using execution feedback.

\item Across different tasks and LLMs, PG consistently outperforms other memory baselines; evolution from scratch produces graphs that match or surpass hand-designed ones, and the loop can also repair flawed expert priors.

\end{itemize}
\section{Related Work}
\label{sec:related_works}
\paragraph{LLM Agents and Action Selection.}
The dominant paradigm follows a free-form action-selection loop: ReAct~\citep{yao2023react} interleaves reasoning with environment actions, and successors extend it with self-critique, branching search, or richer action spaces~\citep{shinn2023reflexion,yao2023tree,wang2024executable,qin2024toolllm}. These approaches rely on the LLM to select valid next actions from in-context information, leaving admissible transitions implicit. Documented failure modes include planning hallucination, drift, and repetitive loops~\citep{zhu2025knowagent,xiao2024flowbench}.

\paragraph{Structured Priors for Agent Planning.} A second line provides explicit structure for planning, including textual procedure rules~\citep{zhu2025knowagent}, workflow knowledge in text, code, or flowchart form~\citep{xiao2024flowbench}, searched workflow graphs~\citep{zhang2025aflow}, and graph-organized tool catalogs~\citep{liu2024toolnet,liu2024controlllm,lumer2025graph}. These methods organize procedural knowledge as action rules, workflows, or tool graphs. PG combines attributed procedure transitions, local retrieval from the current execution context, and refinement of graph topology and attributes.

\paragraph{Self-Improving Agents from Trajectories.} A third line distills reusable knowledge from past trajectories, stored as self-critiques~\citep{shinn2023reflexion}, insights~\citep{zhao2024expel}, state-conditioned guidelines~\citep{fu2024autoguide}, workflows~\citep{wang2025agent}, or procedural memories~\citep{fang2025memp,zhong2024memorybank}. These artifacts retain different forms of structure, including conditional rules and ordered steps within workflows. PG connects transitions across procedure steps in an explicitly editable graph. Its typed, attributed edges support local structural retrieval and refinement from execution feedback. An extended survey and an eight-dimension comparison of 24 methods are provided in Appendix~\ref{app:related_extended}.

\section{The Procedural Graph Framework}
\label{sec:method}

\begin{figure}
    \centering
    \includegraphics[width=\linewidth]{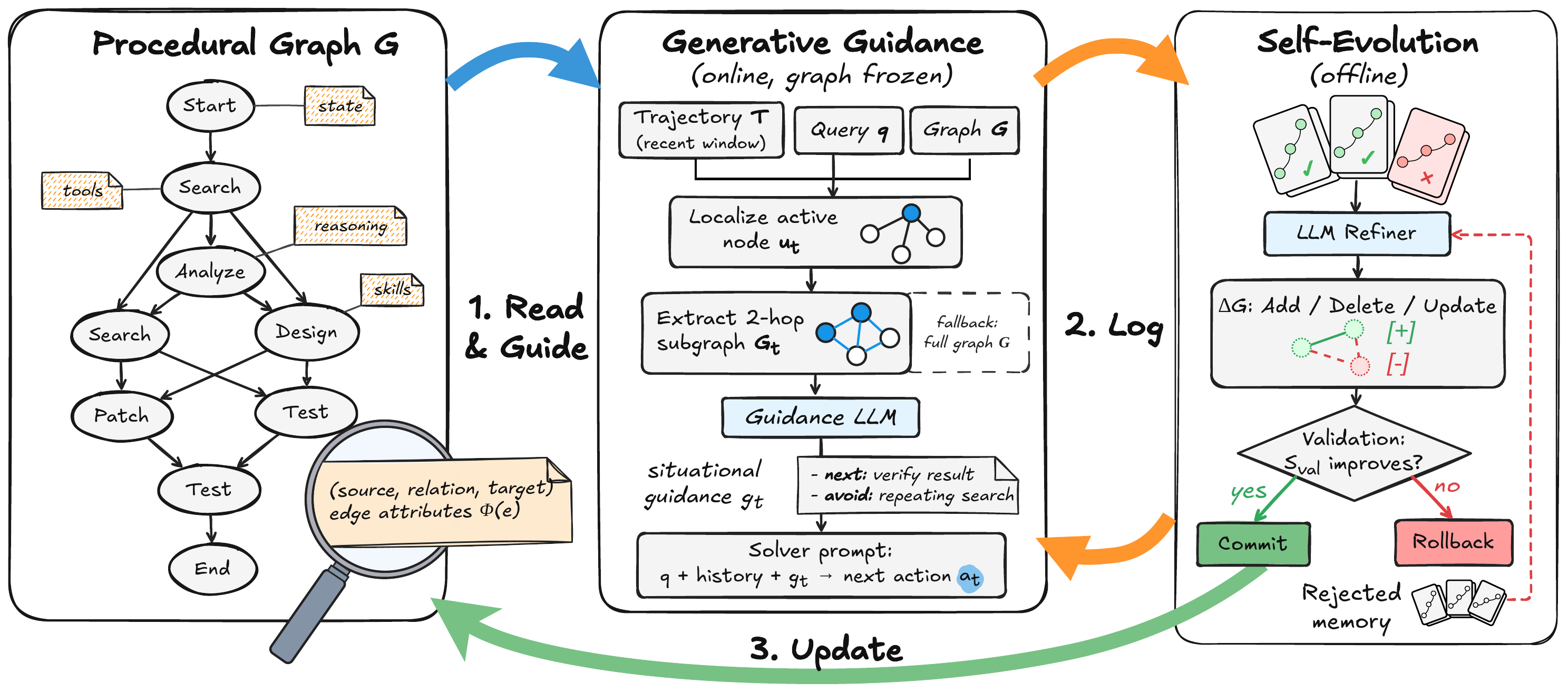}
    \caption{\textbf{Overview of the Procedural Graph framework.} \emph{Left:} procedural triplets define $\mathcal{G}$. \emph{Middle:} the framework localizes $u_t$ and retrieves its $2$-hop neighborhood $\mathcal{G}_t$ (or the full graph if matching fails); a guidance LLM translates it into guidance $g_t$ for the solver. \emph{Right:} the refiner proposes edits from execution trajectories. Structurally valid candidates are committed when validation performance does not decrease; rejected candidates inform subsequent proposals through rejection memory.}
    \label{fig:overall_framework}
\end{figure}

We introduce the \textbf{Procedural Graph (PG)}, a directed graph of procedural knowledge that guides agent execution online while iteratively optimizing its topology and attributes offline. As illustrated in Figure~\ref{fig:overall_framework}, the framework operates in two complementary phases:

\noindent\textbf{Online Inference (Section~\ref{subsec:online_guidance}):} During task solving, the graph is frozen. The agent combines the PG with its trajectory to generate dynamic situational guidance.

\noindent\textbf{Offline Evolution (Section~\ref{subsec:offline_evolution}):} After executing a batch of training tasks, an LLM refiner analyzes the diagnostic traces and modifies the graph topology and attributes via an automated feedback loop.

\subsection{Formal Representation of Procedural Graphs}
\label{subsec:graph_representation}

Formally, a Procedural Graph is a directed, attributed graph
\begin{equation}
    \mathcal{G} = \big( \mathcal{V}, \, \mathcal{R}, \, \mathcal{E}, \, \Phi \big),
    \qquad
    \mathcal{E} \subseteq \mathcal{V} \times \mathcal{R} \times \mathcal{V},
\end{equation}
where $\mathcal{V}$ is the set of abstract nodes, $\mathcal{R}$ is a vocabulary of transition relations, and each element of $\mathcal{E}$ is a directed, attributed triplet: an edge $e = (u, r, v) \in \mathcal{E}$ states that node $v$ is admissible after node $u$ under relation $r$. Each node abstracts a tool function, a skill, an internal reasoning step, or a task status. The attribute mapping $\Phi$ associates each edge with a set of named attributes whose schema can be specified for the task. In our implementation, we use three textual fields: \texttt{condition}, \texttt{guidance}, and \texttt{pitfalls}, describing when the transition applies, how to proceed, and what to avoid. As an illustrative example, a financial-planning edge \texttt{(cash\_flow\_forecast, LEADS\_TO, fund\_raising\_request)} could carry the attributes ``\emph{condition:} projected runway falls below the safety buffer; \emph{guidance:} submit the request early to allow for the financing delivery delay; \emph{pitfalls:} do not stack a second request while one is pending.''

By structuring task knowledge into triplets, $\mathcal{G}$ makes admissible transitions explicit and guides the agent toward valid tool calls and action sequences. The graph can be initialized from an expert prior or from scratch; Section~\ref{sec:pg_construction} compares these construction strategies.

\subsection{Generative Guidance at Inference Time}
\label{subsec:online_guidance}

Independent retrieval of transition attributes, such as top-$k$ similarity search, can omit the connections between procedural steps. For example, retrieving guidance for \texttt{submit} without the preceding \texttt{check\_answer} transition can omit the verification step that makes submission appropriate. Retrieving the connected neighborhood exposes both the action and its procedural prerequisites.

Generative Procedural Graph Guidance addresses this by combining three operations: \emph{locate}, \emph{extract}, and \emph{generate}. Let $q$ be the user query and $\mathcal{T}_t = (a_1, o_1, \dots, a_{t-1}, o_{t-1})$ be the interleaved history of actions and observations up to decision step $t$. We use $a_0 = \texttt{Start}$ as an initialization marker, so the first step is localized at $u_1 = \texttt{Start}$. At each step,
\begin{equation}
    u_t = \mathrm{Match}\big(a_{t-1}, \, \mathcal{V}\big), \qquad
    \mathcal{G}_t =
    \begin{cases}
        \mathcal{N}_h(u_t), & u_t \neq \varnothing, \\
        \mathcal{G}, & \text{otherwise,}
    \end{cases}
    \qquad
    g_t = \Psi\big(\mathcal{G}_t, \; q, \; \mathcal{T}_{t-w:t}\big),
\end{equation}
where $\mathrm{Match}$ locates the agent by exactly matching its most recent procedure (e.g., a tool call) to a node in $\mathcal{V}$. The directed edge neighborhood $\mathcal{N}_h(u_t)$ contains $u_t$ and the outgoing transitions reached by expanding for up to $h$ steps. The window $\mathcal{T}_{t-w:t}$ contains the last $w$ trajectory steps, and $\Psi$ is the guidance language model. This connected neighborhood lets $\Psi$ read transitions in their topological context and consider possible next steps up to $h$ transitions ahead. $\Psi$ translates the static attributes $\Phi(e)$ of the surrounding edges into situational guidance $g_t$, identifying the agent's immediate goal and the formatting or logical errors to avoid.

The guidance $g_t$ is appended to the task solver's prompt. The solver then selects its next action from the query, trajectory, and guidance:
\begin{equation}
    a_t \sim P_{\text{solver}}\big(\cdot \;\big|\; q, \; \mathcal{T}_t, \; g_t\big).
\end{equation}
This soft integration allows the agent to maintain flexible reasoning while being steered toward the procedural structure encoded in $\mathcal{G}$. Guidance uses the localized neighborhood when matching succeeds and the full graph otherwise, together with a recent trajectory window. Section~\ref{subsec:ablation} evaluates the resulting performance and efficiency; Appendix~\ref{app:additional_examples} provides execution cases.

\subsection{Self-Evolution of Procedural Graphs}
\label{subsec:offline_evolution}

An offline self-evolution loop adapts graph topology and attributes from execution feedback, reducing the need for manual design (Algorithm~\ref{alg:evolution}). Let $\mathcal{G}_0$ be the initial Procedural Graph and $\mathcal{G}_k$ the retained graph after round $k$. Each round starts from $\mathcal{G}_{k-1}$; a rejected candidate never becomes the starting graph of the next round. Across generations $k = 1, \dots, K$, the evolution engine executes a four-step loop:

\noindent\textbf{Step 1: Diagnostic Rollout.}
Using the retained graph $\mathcal{G}_{k-1}$, the solver runs on a batch of training tasks $\mathcal{B}_k \subset \mathcal{D}_{\text{train}}$, and we record the diagnostic traces together with their evaluation scores, $\mathcal{E}_k = \big\{ ( q_i, \, \mathcal{T}_i^{(k)}, \, S_i^{(k)} ) \big\}_{i=1}^{|\mathcal{B}_k|}$, where $S_i^{(k)} \in [0, 1]$ is the final task score. The refiner compares high-scoring traces with low-scoring ones; for tasks with binary outcomes, this reduces to successes versus failures.

\noindent\textbf{Step 2: Feedback-Driven Mutation.}
An offline LLM refiner inspects the partitioned traces to identify repeated error loops in failure trajectories and multi-step reasoning shortcuts in successful runs. Based on this feedback, the refiner generates a structured edit set $\Delta\mathcal{G}_k$ comprising two topological edit operations: \textbf{\emph{Add}}: Inserting missing verification nodes or edges; \textbf{\emph{Delete}}: Removing nodes or edges that repeatedly steer trajectories into failure or prevent progress.

Attribute revisions use the same edit interface: an edge is deleted and re-added with updated attribute values. This operation applies to any attribute defined by the chosen schema. The candidate graph is obtained by applying the proposed edits, $\mathcal{G}_k^{\text{cand}} = \mathcal{G}_{k-1} \oplus \Delta\mathcal{G}_k$, where $\oplus$ applies edits to a copy and performs any configured cycle repair.

\noindent\textbf{Step 3: Validation Gating.}
To assess whether structural mutations improve performance beyond the training batch, a candidate that passes edit application and structural checks is evaluated on an independent validation set $\mathcal{D}_{\text{val}}$. Invalid candidates are discarded before validation rollout, leaving the retained graph and its cached validation score unchanged. The mean validation task score is computed as:
\begin{equation}
    S_{\text{val}}(\mathcal{G}) = \frac{1}{|\mathcal{D}_{\text{val}}|} \sum_{(q, y) \in \mathcal{D}_{\text{val}}} S\big( f_{\text{solver}}(q \mid \mathcal{G}), \; y \big).
\end{equation}
The initial graph is evaluated once to establish the reference score. For a structurally valid candidate, the retained graph is updated as follows:
\begin{equation}
    \mathcal{G}_k = 
    \begin{cases} 
        \mathcal{G}_k^{\text{cand}}, & \text{if } S_{\text{val}}(\mathcal{G}_k^{\text{cand}}) \ge S_{\text{val}}(\mathcal{G}_{k-1}) \\ 
        \mathcal{G}_{k-1}, & \text{otherwise} 
    \end{cases}.
\end{equation}
The gate retains candidates whose measured validation score matches or exceeds the cached score of the current graph. Section~\ref{subsec:evolution_results} examines these decisions under stochastic evaluation.

\noindent\textbf{Step 4: Rejection Memory as a Safeguard.}
Iterative self-correction can repeatedly propose equivalent unsuccessful edits. If a candidate graph is rejected by the validation gate ($S_{\text{val}}(\mathcal{G}_k^{\text{cand}}) < S_{\text{val}}(\mathcal{G}_{k-1})$), we log the candidate graph and its proposed edits, together with the associated training trajectories $\mathcal{E}_k$ and validation outcomes, into a rejection memory $\mathcal{H}_{\text{rejected}}$. For the trajectory context supplied to the refiner, we concatenate the training trajectories and, if the configured maximum token length $L_{\max}$ is exceeded, discard tokens from the beginning while preserving the final $L_{\max}$ tokens in their original order. This retains the trajectory ending rather than an initial prefix; the resulting context is denoted as $\mathcal{C}_k$. When proposing edits for round $k+1$, the refiner receives $\mathcal{H}_{\text{rejected}}$ as negative evidence:
\begin{equation}
    \Delta\mathcal{G}_{k+1} \sim P_{\text{refiner}}\big(\cdot \;\big|\; \mathcal{G}_k, \; \mathcal{C}_{k+1}, \; \mathcal{H}_{\text{rejected}}\big).
\end{equation}
The rejection history $\mathcal{H}_{\text{rejected}}$ helps the refiner avoid previously unsuccessful edits. Proposed changes are evaluated by the gate before they enter the retained graph.

\section{Experimental Setup}
\label{sec:experiments}

Implementation details are provided in Appendix~\ref{app:experiment_details}.

\paragraph{Benchmarks.} We evaluate procedural reasoning across seven benchmarks: \textbf{HotpotQA}~\citep{yang2018hotpotqa} for multi-hop question answering with search tools; \textbf{MultiChallenge}~\citep{deshpande2025multichallenge} for instruction retention across multi-turn conversations; \textbf{GDPval}~\citep{patwardhan2025gdpval} for open-ended professional tasks scored against expert rubrics; \textbf{ALFWorld}~\citep{shridharalfworld} for embodied household tasks with strict action ordering; \textbf{$\tau$-bench}~\citep{yao2024tau} for policy-compliant tool use under live user interaction; \textbf{BFCL}~\citep{patil2025berkeley} for multi-turn function calling; and \textbf{EnterpriseArena}~\citep{han2026can} for long-horizon financial decision-making under delayed feedback and macroeconomic shocks. Dataset splits and preprocessing are detailed in Appendix~\ref{app:datasets}, and all metrics are defined in Appendix~\ref{app:metric_details}.

\paragraph{Baselines.} All methods share an identical ReAct solver~\citep{yao2023react} and differ only in how procedural experience is stored and reused; every learning-based baseline consumes the same training trajectories as our self-evolution loop. Ordered by increasing structure, we compare: \textbf{Vanilla ReAct} (no memory)~\citep{yao2023react}, \textbf{MemoryBank}~\citep{zhong2024memorybank}, which maintains summarized experience with forgetting; \textbf{RAP}~\citep{kagaya2024rap}, which retrieves past trajectories as in-context exemplars; \textbf{ExpeL}~\citep{zhao2024expel}, which distills trajectories into natural-language insights; \textbf{AutoGuide}~\citep{fu2024autoguide}, which retrieves state-conditioned guidelines; \textbf{AWM}~\citep{wang2025agent}, which induces linear workflows; and \textbf{KnowAgent}~\citep{zhu2025knowagent}, which maintains textual action-transition rules. Implementation and adaptation details for each baseline are given in Appendix~\ref{app:baseline_details}.

\paragraph{Models.} We evaluate four LLMs: Claude Sonnet 4.6, Gemini 3.1 Pro, Gemini 3.5 Flash, and Grok 4.1 Fast. The guidance model and the offline refiner always share the same underlying LLM as the solver. All calls use greedy decoding (temperature $0$) for reproducibility.

\paragraph{Procedural Graph Configuration.} Online guidance uses the $h{=}2$ hop neighborhood of the localized node and a recent trajectory window of $w{=}3$. Different construction strategies are compared in Section~\ref{sec:pg_construction} and formalized in Appendix~\ref{app:mode_details}. Statistics of the graphs used for each benchmark (node and triplet counts, relation types, attribute coverage) are given in Appendix~\ref{app:pg_stats}.

\section{Results}

\subsection{Main Results across Benchmarks and Models}

Table~\ref{tab:llm_six_datasets} compares the Procedural Graph against seven baselines across six benchmarks and four LLM families, all using the same ReAct solver. PG ranks first or joint first in 21 of 24 model--benchmark settings. Compared with the strongest baseline in each setting, PG records 19 wins, two ties, and three losses (one-sided exact binomial sign test excluding ties, $p = 4.3\times10^{-4}$). Its largest margins are on BFCL v3 with Gemini 3.5 Flash ($67.00\%$ vs.\ $58.00\%$, $+9.00$ points), GDPval with Gemini 3.1 Pro ($78.78$ vs.\ $71.37$, $+7.41$ points), and $\tau$-bench with the same model ($80.00\%$ vs.\ $73.04\%$, $+6.96$ points). Baseline rankings vary across tasks and models, with no single method consistently placing second. These results suggest that combining conditional guidance, reusable action sequences, and explicit transitions in a connected graph is useful across diverse settings.

\begin{table*}[thbp]
\centering
\scriptsize
\newcommand{\ci}[2]{#1\,{\fontsize{4pt}{6pt}\selectfont [#2]}}
\newcommand{\cib}[2]{\cellcolor{GoogleGreen}\textcolor{white}{\textbf{#1}\,{\fontsize{4pt}{6pt}\selectfont \textbf{[#2]}}}}
\setlength{\aboverulesep}{0pt}
\setlength{\belowrulesep}{0pt}
\renewcommand{\arraystretch}{1}
\setlength{\tabcolsep}{1.5pt}
\caption{\textbf{Main results across LLMs and benchmarks.} Brackets give $95\%$ confidence intervals; the best value is highlighted.}
\label{tab:llm_six_datasets}
\begin{tabular}{l c c c c c c}
\toprule
\rowcolor{GoogleGreen!15}
\cellcolor{GoogleGreen!15} & \textbf{HotpotQA} & \textbf{MultiChallenge} & \textbf{GDPval} & \textbf{ALFWorld} & \textbf{$\tau$-bench} & \textbf{BFCL v3} \\

\arrayrulecolor{white}
\cmidrule(lr){2-2} \cmidrule(lr){3-3} \cmidrule(lr){4-4} \cmidrule(lr){5-5} \cmidrule(lr){6-6} \cmidrule(lr){7-7}
\arrayrulecolor{black}

\rowcolor{GoogleGreen!15}
\multirow{-2}{*}{\textbf{Model \& Method}} & \textbf{Acc. ($\uparrow$)} & \textbf{Acc. ($\uparrow$)} & \textbf{Rubric Score ($\uparrow$)} & \textbf{Success ($\uparrow$)} & \textbf{Pass@1 ($\uparrow$)} & \textbf{Acc. ($\uparrow$)} \\
\midrule

\multicolumn{7}{l}{\textbf{Claude Sonnet 4.6}} \\
\midrule
\quad Vanilla ReAct~\citep{yao2023react}       & \ci{74.60}{71.83, 77.37} & \ci{83.73}{78.06, 89.41} & \ci{46.23}{33.42, 59.04} & \ci{86.57}{80.76, 92.37} & \ci{66.09}{57.54, 74.64} & \ci{61.00}{51.29, 70.71} \\
\quad MemoryBank~\citep{zhong2024memorybank}   & \ci{74.20}{71.50, 76.90} & \ci{89.16}{84.50, 93.82} & \ci{50.13}{37.04, 63.22} & \ci{86.57}{80.72, 92.41} & \ci{63.48}{54.78, 72.18} & \ci{62.00}{52.55, 71.45} \\
\quad RAP~\citep{kagaya2024rap}                & \ci{71.80}{69.07, 74.53} & \ci{86.75}{81.56, 91.93} & \ci{40.97}{28.37, 53.56} & \ci{79.10}{72.28, 85.93} & \ci{60.87}{52.12, 69.62} & \ci{65.00}{55.68, 74.32} \\
\quad ExpeL~\citep{zhao2024expel}              & \ci{75.20}{72.49, 77.91} & \ci{89.16}{84.45, 93.86} & \ci{48.22}{35.66, 60.78} & \ci{91.79}{87.30, 96.28} & \ci{71.30}{63.05, 79.56} & \ci{60.00}{50.46, 69.54} \\
\quad AutoGuide~\citep{fu2024autoguide}        & \cib{75.40}{72.72, 78.08} & \ci{88.55}{83.71, 93.40} & \ci{41.39}{27.57, 55.21} & \ci{84.33}{78.12, 90.54} & \ci{65.22}{56.54, 73.89} & \ci{59.00}{49.19, 68.81} \\
\quad AWM~\citep{wang2025agent}                & \ci{73.90}{71.21, 76.59} & \cib{89.76}{85.17, 94.34} & \ci{40.61}{27.54, 53.68} & \ci{67.16}{59.23, 75.10} & \ci{66.96}{58.51, 75.40} & \ci{62.00}{52.56, 71.44} \\
\quad KnowAgent~\citep{zhu2025knowagent}       & \ci{74.30}{71.59, 77.01} & \ci{87.35}{82.36, 92.34} & \ci{42.14}{29.50, 54.78} & \ci{87.31}{81.58, 93.05} & \ci{61.74}{52.67, 70.81} & \ci{60.00}{50.29, 69.71} \\
\quad Procedural Graph (Ours)                  & \ci{74.50}{71.81, 77.19} & \cib{89.76}{85.21, 94.31} & \cib{51.49}{38.44, 64.54} & \cib{93.28}{88.98, 97.59} & \cib{73.91}{65.84, 81.99} & \cib{67.00}{57.65, 76.35} \\
\midrule

\multicolumn{7}{l}{\textbf{Gemini 3.1 Pro}} \\
\midrule
\quad Vanilla ReAct~\citep{yao2023react}       & \ci{85.90}{83.76, 88.04} & \ci{87.95}{83.00, 92.91} & \ci{56.39}{45.23, 67.55} & \ci{94.78}{91.03, 98.52} & \ci{72.17}{63.92, 80.43} & \ci{59.00}{49.33, 68.67} \\
\quad MemoryBank~\citep{zhong2024memorybank}   & \ci{85.10}{82.90, 87.30} & \ci{95.18}{92.00, 98.36} & \ci{61.97}{50.54, 73.39} & \ci{85.07}{78.87, 91.28} & \ci{67.83}{59.38, 76.27} & \ci{63.00}{53.69, 72.31} \\
\quad RAP~\citep{kagaya2024rap}                & \ci{83.60}{81.28, 85.92} & \ci{94.58}{91.17, 97.99} & \ci{71.37}{62.99, 79.76} & \ci{80.60}{73.87, 87.32} & \ci{73.04}{64.98, 81.11} & \ci{57.00}{47.20, 66.80} \\
\quad ExpeL~\citep{zhao2024expel}              & \ci{86.00}{83.85, 88.15} & \ci{92.77}{88.84, 96.70} & \ci{64.69}{54.33, 75.06} & \ci{97.76}{95.25, 100.00} & \ci{64.35}{55.46, 73.24} & \ci{63.00}{53.60, 72.40} \\
\quad AutoGuide~\citep{fu2024autoguide}        & \ci{85.80}{83.63, 87.97} & \ci{93.37}{89.54, 97.21} & \ci{56.18}{44.73, 67.62} & \ci{91.04}{86.17, 95.92} & \ci{65.22}{56.59, 73.84} & \ci{63.00}{53.60, 72.40} \\
\quad AWM~\citep{wang2025agent}                & \ci{85.10}{82.89, 87.31} & \ci{93.98}{90.45, 97.50} & \ci{63.85}{54.13, 73.57} & \ci{99.25}{97.78, 100.00} & \ci{67.83}{59.32, 76.33} & \ci{64.00}{54.41, 73.59} \\
\quad KnowAgent~\citep{zhu2025knowagent}       & \ci{85.00}{82.82, 87.18} & \ci{95.18}{91.89, 98.47} & \ci{69.10}{60.25, 77.95} & \ci{95.52}{91.99, 99.06} & \ci{64.35}{55.42, 73.28} & \ci{61.00}{51.44, 70.56} \\
\quad Procedural Graph (Ours)                  & \cib{87.30}{85.18, 89.42} & \cib{95.78}{92.78, 98.79} & \cib{78.78}{74.42, 83.15} & \cib{100.00}{100.00, 100.00} & \cib{80.00}{72.83, 87.17} & \cib{66.00}{56.85, 75.15} \\
\midrule

\multicolumn{7}{l}{\textbf{Gemini 3.5 Flash}} \\
\midrule
\quad Vanilla ReAct~\citep{yao2023react}       & \ci{83.10}{80.79, 85.41} & \ci{81.33}{75.33, 87.33} & \ci{59.33}{47.49, 71.16} & \ci{82.84}{76.42, 89.25} & \ci{31.30}{22.77, 39.84} & \ci{56.00}{46.51, 65.49} \\
\quad MemoryBank~\citep{zhong2024memorybank}   & \ci{82.50}{80.14, 84.86} & \ci{89.16}{84.33, 93.99} & \ci{59.27}{47.81, 70.72} & \ci{76.87}{69.89, 83.84} & \ci{34.78}{26.03, 43.53} & \ci{54.00}{44.24, 63.76} \\
\quad RAP~\citep{kagaya2024rap}                & \ci{82.30}{79.97, 84.63} & \ci{89.16}{84.45, 93.86} & \ci{60.23}{48.63, 71.82} & \ci{76.12}{69.00, 83.24} & \ci{37.39}{28.45, 46.34} & \ci{54.00}{44.11, 63.89} \\
\quad ExpeL~\citep{zhao2024expel}              & \ci{83.40}{81.09, 85.71} & \ci{89.16}{84.42, 93.89} & \ci{62.45}{50.97, 73.93} & \ci{90.30}{85.34, 95.26} & \ci{32.17}{23.61, 40.73} & \ci{58.00}{48.50, 67.50} \\
\quad AutoGuide~\citep{fu2024autoguide}        & \ci{83.10}{80.82, 85.38} & \ci{89.16}{84.32, 93.99} & \ci{50.29}{37.54, 63.04} & \ci{80.60}{73.87, 87.32} & \ci{26.09}{18.08, 34.10} & \ci{56.00}{46.01, 65.99} \\
\quad AWM~\citep{wang2025agent}                & \ci{82.50}{80.14, 84.86} & \cib{91.57}{87.38, 95.75} & \ci{50.25}{37.54, 62.95} & \ci{82.09}{75.55, 88.63} & \ci{33.91}{25.37, 42.46} & \ci{52.00}{42.24, 61.76} \\
\quad KnowAgent~\citep{zhu2025knowagent}       & \ci{82.70}{80.39, 85.01} & \cib{91.57}{87.26, 95.88} & \ci{54.00}{41.68, 66.31} & \ci{81.34}{74.86, 87.83} & \ci{38.26}{29.23, 47.29} & \ci{58.00}{48.31, 67.69} \\
\quad Procedural Graph (Ours)                  & \cib{84.50}{82.30, 86.70} & \cib{91.57}{87.33, 95.80} & \cib{64.42}{53.66, 75.19} & \cib{94.03}{90.03, 98.03} & \cib{44.35}{35.20, 53.50} & \cib{67.00}{57.89, 76.11} \\
\midrule

\multicolumn{7}{l}{\textbf{Grok 4.1 Fast}} \\
\midrule
\quad Vanilla ReAct~\citep{yao2023react}       & \ci{72.70}{69.89, 75.51} & \ci{68.07}{61.04, 75.10} & \ci{57.23}{47.11, 67.34} & \ci{26.12}{18.64, 33.60} & \ci{64.35}{55.58, 73.11} & \ci{52.00}{42.29, 61.71} \\
\quad MemoryBank~\citep{zhong2024memorybank}   & \ci{70.70}{67.84, 73.56} & \ci{84.94}{79.47, 90.41} & \ci{55.19}{44.40, 65.98} & \ci{23.88}{16.79, 30.97} & \ci{58.26}{49.48, 67.04} & \ci{56.00}{46.10, 65.90} \\
\quad RAP~\citep{kagaya2024rap}                & \ci{70.20}{67.32, 73.08} & \ci{80.12}{74.01, 86.24} & \ci{66.79}{58.25, 75.33} & \ci{14.93}{8.90, 20.95} & \ci{64.35}{55.41, 73.28} & \ci{55.00}{45.22, 64.78} \\
\quad ExpeL~\citep{zhao2024expel}              & \ci{73.50}{70.81, 76.19} & \ci{84.94}{79.53, 90.35} & \ci{65.66}{57.00, 74.32} & \cib{42.54}{34.08, 50.99} & \ci{63.48}{54.72, 72.24} & \ci{52.00}{42.36, 61.64} \\
\quad AutoGuide~\citep{fu2024autoguide}        & \ci{70.00}{67.15, 72.85} & \ci{80.72}{74.73, 86.72} & \ci{53.58}{42.90, 64.26} & \ci{30.60}{22.74, 38.45} & \ci{61.74}{52.90, 70.57} & \ci{56.00}{45.99, 66.01} \\
\quad AWM~\citep{wang2025agent}                & \ci{74.30}{71.60, 77.00} & \ci{83.73}{78.26, 89.21} & \ci{61.55}{52.66, 70.44} & \ci{17.16}{10.84, 23.49} & \ci{63.48}{54.87, 72.09} & \ci{52.00}{42.20, 61.80} \\
\quad KnowAgent~\citep{zhu2025knowagent}       & \ci{73.10}{70.34, 75.86} & \ci{83.13}{77.55, 88.72} & \ci{68.15}{61.15, 75.16} & \ci{18.66}{12.11, 25.20} & \cib{68.70}{60.45, 76.94} & \ci{54.00}{44.44, 63.56} \\
\quad Procedural Graph (Ours)                  & \cib{74.40}{71.62, 77.18} & \cib{86.75}{81.74, 91.76} & \cib{71.19}{65.24, 77.14} & \ci{39.55}{31.11, 47.99} & \ci{67.83}{59.27, 76.39} & \cib{60.00}{50.51, 69.49} \\
\bottomrule
\end{tabular}
\end{table*}

The gains also extend across model families. On GDPval and BFCL v3, PG outperforms every baseline under all four LLMs, indicating that its advantage on these tasks is not confined to a particular solver. On MultiChallenge, PG ranks first or joint first across all four models, matching AWM on Claude Sonnet 4.6 and both AWM and KnowAgent on Gemini 3.5 Flash. HotpotQA shows a different pattern: margins over the strongest baseline range from $-0.90$ to $+1.30$ points. The magnitude of the gains therefore varies substantially across benchmarks.

\subsection{Long-Horizon Decision Making and Resilience}
\label{subsec:long_horizon}

\begin{figure}
    \centering
    \includegraphics[width=\linewidth]{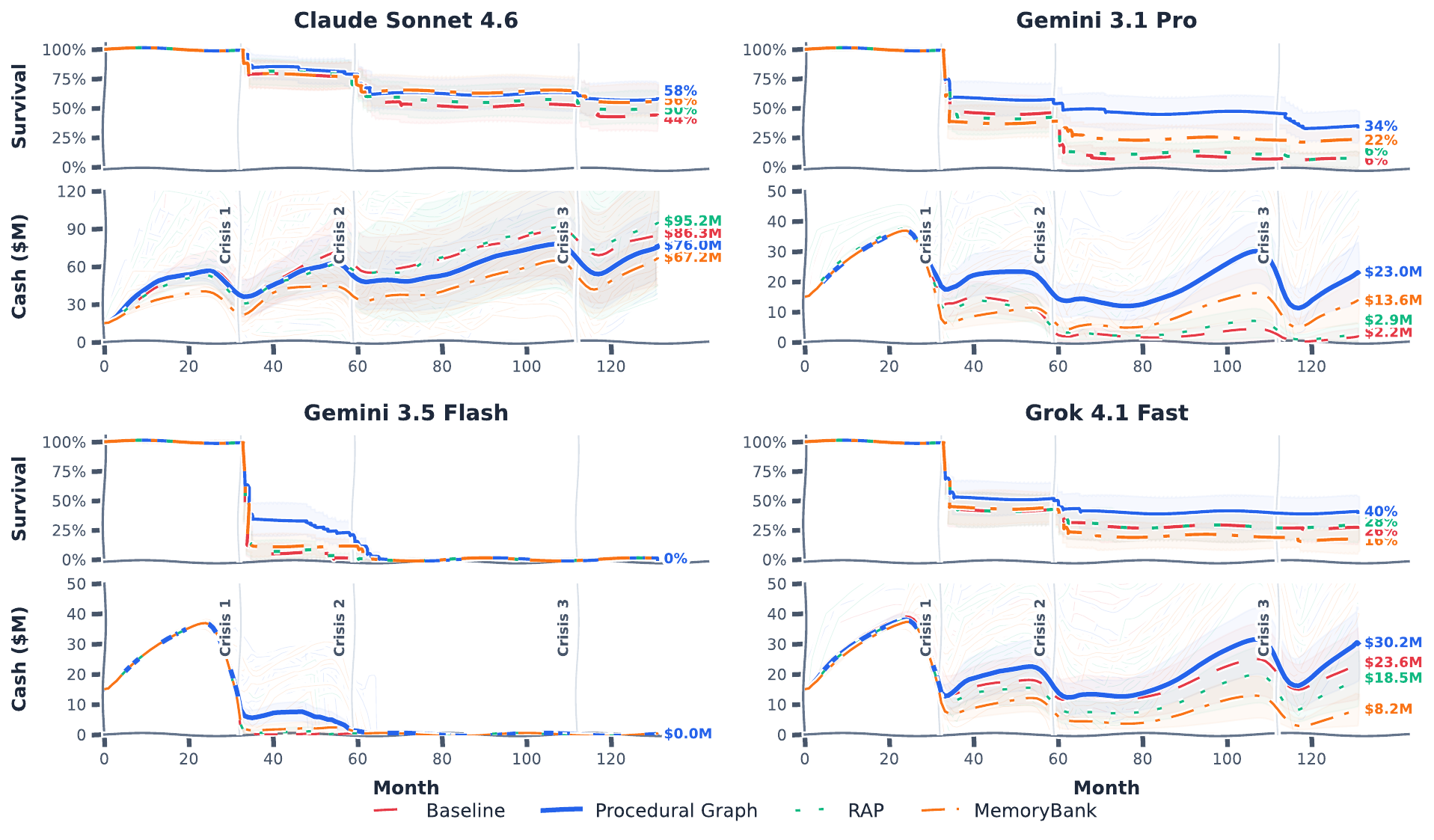}
    \caption{\textbf{Ensemble cash trajectories and Kaplan--Meier survival curves across four LLMs.} Bold lines and shaded areas show means and $95\%$ confidence intervals. Vertical lines mark macroeconomic crises. Colors identify PG (blue), the baseline (red), and memory-based methods (green, orange).}
    \label{fig:cfo_ensemble_trajectories}
\end{figure}

To evaluate agent resilience on long-horizon tasks, we deploy agents in EnterpriseArena~\citep{han2026can}, a simulator in which the agent makes monthly financial decisions over up to 132 months under strict liquidity constraints and three scheduled crises that are not disclosed to the agent. Figure~\ref{fig:cfo_ensemble_trajectories} plots the Kaplan-Meier survival curves and the ensemble cash trajectories for all four models; complete metrics are reported in Table~\ref{tab:gemini_grok_simulation} in Appendix~\ref{app:cfo_appendix}. PG achieves the highest or joint-highest full-horizon survival and the longest average lifespan across all four LLMs. It raises survival from $44.0\%$ to $58.0\%$ for Claude Sonnet 4.6, from $6.0\%$ to $34.0\%$ for Gemini 3.1 Pro, and from $26.0\%$ to $40.0\%$ for Grok 4.1 Fast, where it also delivers the best average enterprise score ($\$39.62$M).

What changes under guidance is \emph{which} tools are called and when, rather than simply how many. The unguided Gemini 3.5 Flash baseline repeatedly queries cash and market state within a single turn, adding redundant observations to its context. It issues $18.94$ tool calls per month, which the graph reduces to $12.53$ while improving the average enterprise score. On Claude Sonnet 4.6 and Gemini 3.1 Pro, tool calls instead increase from $0.13$ to $0.36$ and from $0.89$ to $3.18$ per month, respectively, while survival also improves. In these settings, the graph guides the agent to run forecast and market checks before a financing decision (Appendix~\ref{app:cfo_plots}).

The behavior that does track survival across all four models is anticipatory fundraising. Because capital arrives one to six months after it is requested, surviving a crisis requires asking well before liquidity runs out. The full trajectories show that the unguided Gemini 3.5 Flash baseline does not initiate fundraising sufficiently early, whereas PG-guided agents initiate requests during stable months. Average capital raised is $\$0.00$M for the Flash baseline, compared with $\$9.39$M for PG-guided Flash and $\$30.11$M for PG-guided Grok 4.1 Fast. Appendix~\ref{app:cfo_traces} contrasts step-by-step traces of the unguided baseline, a memory-summarization agent, and a PG-guided agent entering the first crisis.

\subsection{Procedural Graph Construction Strategies}
\label{sec:pg_construction}
We compare five PG construction strategies against the unguided baseline, spanning expert versus minimal initialization and fixed, one-time, or iterative refinement. Modes 3 and 5 instantiate the full evolution loop, while Modes 1, 2, and 4 provide fixed or one-time alternatives (supplementary results and mode definitions are provided in Appendix~\ref{app:procedural_graph_construction}). The performance metrics of HotpotQA and MultiChallenge are presented in Table~\ref{tab:performance_metrics}.

\begin{table*}[thbp]
\scriptsize
\centering
\setlength{\aboverulesep}{0pt}
\setlength{\belowrulesep}{0pt}
\renewcommand{\arraystretch}{1}
\setlength{\tabcolsep}{5.4pt}
\caption{\textbf{PG construction modes on HotpotQA and MultiChallenge.} All metrics are higher-is-better.}
\label{tab:performance_metrics}
\begin{tabular}{l c c c c c c c}
\toprule
\rowcolor{GoogleBlue!15}
\cellcolor{GoogleBlue!15} & \multicolumn{2}{c}{\textbf{HotpotQA}} & \multicolumn{5}{c}{\textbf{MultiChallenge}} \\
\arrayrulecolor{white}\cmidrule(lr){2-3} \cmidrule(lr){4-8}\arrayrulecolor{black}
\rowcolor{GoogleBlue!15}
\cellcolor{GoogleBlue!15} & \cellcolor{GoogleBlue!15} & \cellcolor{GoogleBlue!15} & \textbf{Inference} & \textbf{Instruction} & \textbf{Reliable} & \textbf{Self} & \cellcolor{GoogleBlue!15} \\
\rowcolor{GoogleBlue!15}
\multirow{-3}{*}{\textbf{Construction Mode}} & \multirow{-2}{*}{\textbf{Ans EM}} & \multirow{-2}{*}{\textbf{Ans F1}} & \textbf{Memory} & \textbf{Retention} & \textbf{Versioned Editing} & \textbf{Coherence} & \multirow{-3}{*}{\textbf{Overall}} \\
\midrule
Unguided Baseline (w/o PG)         & 58.8 & 71.21 & 86.96 & 86.67 & 71.43 & \cellcolor{GoogleBlue}\textcolor{white}{\textbf{100.00}} & 87.50 \\
Mode 1: Hand-crafted Expert        & 62.80 & 76.61 & 52.17 & 66.67 & 42.86 & 72.73 & 58.93 \\
Mode 2: Expert + Static Update     & 63.80 & 77.16 & 47.83 & 53.33 & 28.57 & 81.82 & 53.57 \\
Mode 3: Expert + Online Evolution  & 63.10 & 76.34 & \cellcolor{GoogleBlue}\textcolor{white}{\textbf{95.65}} & \cellcolor{GoogleBlue}\textcolor{white}{\textbf{100.00}} & \cellcolor{GoogleBlue}\textcolor{white}{\textbf{85.71}} & 81.82 & \cellcolor{GoogleBlue}\textcolor{white}{\textbf{92.86}} \\
Mode 4: Scratch + Static Build     & 55.40 & 69.49 & 91.30 & 86.67 & \cellcolor{GoogleBlue}\textcolor{white}{\textbf{85.71}} & 90.91 & 89.29 \\
Mode 5: Scratch + Online Evolution & \cellcolor{GoogleBlue}\textcolor{white}{\textbf{66.30}} & \cellcolor{GoogleBlue}\textcolor{white}{\textbf{78.79}} & \cellcolor{GoogleBlue}\textcolor{white}{\textbf{95.65}} & 93.33 & 71.43 & 90.91 & 91.07 \\
\bottomrule
\end{tabular}
\end{table*}
We compare initialization from an expert prior (Modes 1--3) with initialization from scratch (Modes 4--5). On HotpotQA, Mode 5 (Scratch + Online Evolution) achieves the highest performance across all configurations, scoring $78.79\%$ Ans F1 and $66.30\%$ Ans EM (gains of $7.58$ F1 points and $7.50$ EM points over the unguided baseline). On MultiChallenge, which requires retaining multiple constraints across dialogue turns, Mode 5 achieves an Overall Success Rate of $91.07\%$ without a human prior, while Mode 3 (Expert + Online Evolution) performs best at $92.86\%$.

The loop also self-corrects from a flawed expert prior. On MultiChallenge, using the hand-crafted expert graph (Mode 1) lowers success from $87.50\%$ to $58.93\%$. A single offline update (Mode 2) further lowers success to $53.57\%$. The iterative configuration (Mode 3), which combines fresh execution feedback with validation gating, recovers to $92.86\%$, a gain of $33.93$ points over the expert initialization. The loop therefore recovers from an expert prior that initially reduces performance (Appendix~\ref{app:multichallenge_edits}). Resource and stability statistics for all five modes, including a $45.7\%$ reduction in parsing failures and the token overhead carried by expert-written guidance, are reported in Appendix~\ref{app:efficiency}.

\subsection{Procedural Graph Self-Evolution}
\label{subsec:evolution_results}

\begin{figure}[htbp]
    \centering
    \includegraphics[width=\linewidth]{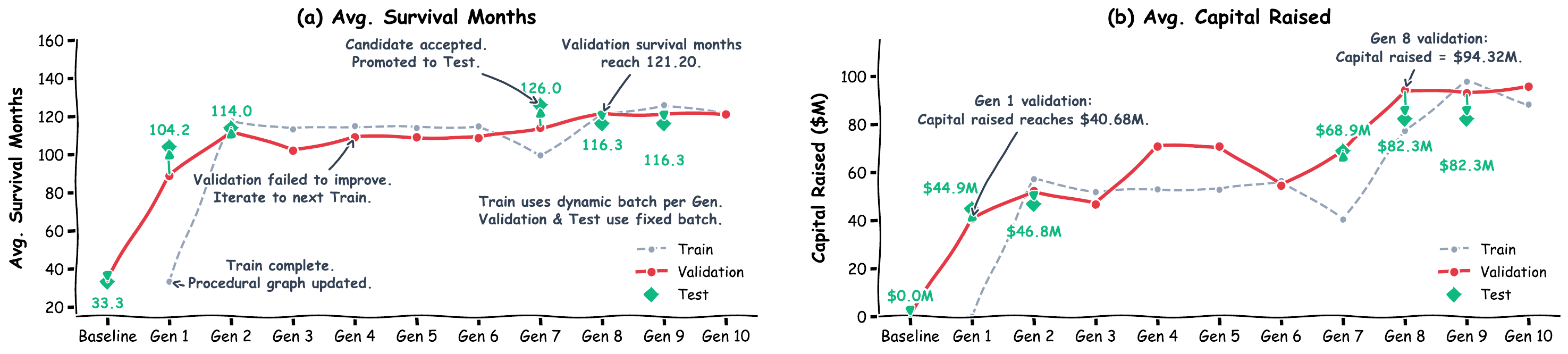}
    \caption{\textbf{Mean lifespan and capital raised across ten rounds of PG self-evolution.} Gray dashed lines show training results; red lines show validation results; and green diamonds show test results for the baseline and accepted checkpoints.}
    \label{fig:loop_evolution}
\end{figure}

We examine ten rounds of PG self-evolution (Section~\ref{subsec:offline_evolution}) on EnterpriseArena, where the agent manages liquidity through successive macroeconomic crises. Figure~\ref{fig:loop_evolution} tracks the resulting changes in lifespan and capital raised; Appendix~\ref{app:self_evolution} reports the per-round results.

On the validation split, the unguided baseline has a full-horizon survival rate of $0.0\%$ and a mean lifespan of $34.8$ months. Since capital takes one to six months to arrive, fundraising must begin before cash runs out. Validation performance improves in several distinct rounds, separated by periods with no accepted update. Round 1 discovers the sequential backbone that guides the agent to audit cash and forecast runway before a financing decision, producing the largest single jump ($0.0\% \rightarrow 45.0\%$ validation survival); Round 2 adds \texttt{recall\_notes} to reuse the notes saved by the Round 1 graph and lifts survival to $80.0\%$, with tool usage falling from $17.23$ to $3.08$ calls per month relative to the unguided baseline. Rounds 3--6 produce no committed update; one candidate fails structural verification before rollout. Round 7 prunes the \texttt{pass\_action} branch, and Round 8 introduces the administrative bypass described in Appendix~\ref{sec:app_topology}. Validation survival reaches $90.0\%$ in Round 8 and remains at that level in Round 9 before Round 10 is rejected and the loop terminates.

The test results distinguish the returned graph from intermediate candidates. The returned graph reaches $85.0\%$ test survival against the baseline's $0.0\%$ (Fisher's exact $p = 2.6\times10^{-8}$), while the best single round observed during the search reached $95.0\%$; we report the former, since quoting the latter would amount to selecting on the test set. With $20$ episodes per split, individual accept/reject decisions turn on one or two episodes and should be read as a search trace rather than as significance tests. Topological changes and cross-dataset validation are detailed in Appendices~\ref{sec:app_topology} and~\ref{sec:app_cross_validation}.

\subsection{Efficiency Analysis}
\label{subsec:ablation}

Table~\ref{tab:ablation_graph_configs} examines two key design choices behind our guidance mechanism: what portion of the graph the agent sees (full graph vs. localized subgraph) and how it is consumed (raw injection vs. generative guidance). The three PG configurations compare generative guidance with raw injection for the full graph and assess localization under generative guidance, alongside a no-graph baseline. All configurations use Gemini 3.5 Flash and a shared solver prompt template; the PG configurations use the same underlying graph (Appendix~\ref{app:prompts}).

\begin{table*}[htbp]
\centering
\scriptsize
\setlength{\aboverulesep}{0pt}
\setlength{\belowrulesep}{0pt}
\renewcommand{\arraystretch}{1}
\setlength{\tabcolsep}{4.7pt}
\caption{\textbf{PG usage ablation with Gemini 3.5 Flash.} On fixed subsets, we report MultiChallenge accuracy, GDPval rubric score, and ALFWorld success rate (all $\uparrow$), alongside average tokens and solver steps per sample (both $\downarrow$).}
\label{tab:ablation_graph_configs}
\begin{tabular}{l ccc ccc ccc}
\toprule
\rowcolor{GoogleYellow!15}
 & \multicolumn{3}{c}{\textbf{MultiChallenge}} & \multicolumn{3}{c}{\textbf{GDPval}} & \multicolumn{3}{c}{\textbf{ALFWorld}} \\
\arrayrulecolor{white}\cmidrule(lr){2-4}\cmidrule(lr){5-7}\cmidrule(lr){8-10}\arrayrulecolor{black}
\rowcolor{GoogleYellow!15}
\multirow{-2}{*}{\textbf{Graph Configuration}} & \textbf{Acc.} & \textbf{Avg. Tok} & \textbf{Avg. Step} & \textbf{Rubric} & \textbf{Avg. Tok} & \textbf{Avg. Step} & \textbf{Success} & \textbf{Avg. Tok} & \textbf{Avg. Step} \\
\midrule
Baseline (no graph)            & 80.27 & 6{,}629 & 3.87 & 54.80 & 275{,}638 & 28.20 & 72.58 & 18{,}055 & 21.84 \\
Full graph, raw injection      & 86.60 & 10{,}164 & 4.54 & 57.17 & 264{,}680 & 33.55 & 70.34 & 21{,}062 & 25.00 \\
Full graph, generative         & 87.35 & 14{,}434 & \cellcolor{GoogleYellow}\textcolor{black}{\textbf{3.08}} & 56.75 & 448{,}972 & 22.07 & 54.48 & 96{,}360 & 30.05 \\
Subgraph, generative (Ours)    & \cellcolor{GoogleYellow}\textcolor{black}{\textbf{89.31}} & 12{,}295 & 4.22 & \cellcolor{GoogleYellow}\textcolor{black}{\textbf{63.99}} & 367{,}738 & \cellcolor{GoogleYellow}\textcolor{black}{\textbf{18.57}} & \cellcolor{GoogleYellow}\textcolor{black}{\textbf{81.53}} & 28{,}064 & \cellcolor{GoogleYellow}\textcolor{black}{\textbf{18.80}} \\
\bottomrule
\end{tabular}
\end{table*}

Injecting the raw full graph improves performance on structured dialogue (MultiChallenge rises from $80.27$ to $86.60$) but lowers success on embodied execution (ALFWorld drops from $72.58$ to $70.34$). Full-graph generative guidance further reduces ALFWorld success to $54.48$ while increasing token consumption. These results favor guidance grounded in the agent's local graph neighborhood over guidance generated from the full graph. Given the same graph, the localized generative configuration achieves the highest performance across all three benchmarks ($89.31$, $63.99$, and $81.53$), exceeding the best alternative in each setting, including the no-graph baseline, by $2.0$, $6.8$, and $9.0$ points, respectively.

Localization reduces total tokens relative to full-graph generative guidance on all three benchmarks: by $70.9\%$ on ALFWorld, $18.1\%$ on GDPval, and $14.8\%$ on MultiChallenge. It also shortens trajectories on GDPval and ALFWorld. The additional guidance call introduces token overhead relative to the no-graph baseline. On GDPval and ALFWorld, localized guidance reduces average solver steps from $28.20$ to $18.57$ and from $21.84$ to $18.80$, respectively, while total token consumption remains $33.4\%$ and $55.4\%$ higher.

\section{Conclusion}

We introduced the Procedural Graph, an explicit and editable representation of procedural knowledge that gives LLM agents a queryable answer to what to do next. PG connects the agent's current progress with relevant transitions and execution advice while preserving reasoning flexibility. Across tasks and model families, it delivers consistent gains over memory-based baselines. Self-evolution builds effective graphs from minimal initializations and repairs expert priors that initially hinder performance. These results support learning and revising procedural knowledge from execution feedback without updating model weights. Guidance increases token use even when it reduces solver steps; future work could reuse guidance across steps or generate it selectively. Evaluating transfer across solvers and tool interfaces would clarify how widely the learned procedures can be reused.

\clearpage

\appendix

\section{Extended Related Work and Comparison}
\label{app:related_extended}

This appendix expands Section~\ref{sec:related_works}. It covers procedural memory (Appendix~\ref{app:coala}), action selection (Appendix~\ref{app:agents-extended}), structured planning (Appendix~\ref{app:priors-extended}), and self-improvement from trajectories (Appendix~\ref{app:trajectory-extended}). A comparison along eight design dimensions appears in Appendix~\ref{app:comparison}.

\subsection{Procedural Graphs as Procedural Memory: A CoALA View}
\label{app:coala}

\citet{sumers2023cognitive} propose CoALA (Cognitive Architectures for Language Agents), which adapts the classical memory taxonomy of ACT-R and SOAR to LLM-based agents: \emph{working memory} holds the current context, \emph{episodic memory} holds past experiences, \emph{semantic memory} holds factual knowledge, and \emph{procedural memory} holds the skills and procedures that govern how to act. This taxonomy distinguishes the roles of different agent memories. Retrieval augmentation, including GraphRAG and its variants, operates on semantic memory; trajectory-based reflection methods such as Reflexion~\citep{shinn2023reflexion} and ExpeL~\citep{zhao2024expel} operate on episodic memory. Procedural memory is the quadrant that has received the least explicit treatment: it remains largely implicit in model weights, or is scattered across ad hoc artifacts such as prompt templates, skill libraries, and workflow scripts. PG implements CoALA's procedural-memory module by storing state-conditioned action transitions in a structure that can be retrieved and updated.

\subsection{LLM Agents and Action Selection}
\label{app:agents-extended}

The dominant execution template is reason-then-act interleaving, established by ReAct~\citep{yao2023react} and extended along several axes. Reflexion~\citep{shinn2023reflexion} inserts verbal self-criticism between trials; Toolformer~\citep{schick2023toolformer} shows that tool-call decisions can be learned by self-supervised filtering of LM-generated API calls; Tree-of-Thoughts~\citep{yao2023tree} and Graph-of-Thoughts~\citep{besta2024graph} generalize chain-style reasoning into search over branching alternatives; and Plan-and-Solve~\citep{wang2023plan} and ADaPT~\citep{prasad2024adapt} separate planning from execution, with the latter recursively decomposing sub-tasks only once the executor fails. A parallel line redesigns the action space itself rather than the control flow over it: CodeAct~\citep{wang2024executable} unifies actions as executable Python so as to inherit the compositionality of a programming language, and HuggingGPT~\citep{shen2023hugginggpt} treats models hosted on Hugging Face as callable tools coordinated by an LLM controller.

A third body of work scales the tool catalog. ToolLLM~\citep{qin2024toolllm} contributes a 16{,}464-API benchmark together with a depth-first decision tree for tool selection; Gorilla~\citep{patil2024gorilla} fine-tunes LLaMA on APIBench with retrieval-augmented training; AnyTool~\citep{du2024anytool} adds a hierarchical three-tier API retriever; and ToolGen~\citep{wang2025toolgen} collapses retrieval and invocation into next-token generation via virtual tool tokens. API-Bank~\citep{li2023api} and MetaTool~\citep{huang2024metatool} supply complementary evaluation axes covering when to invoke a tool and which one to invoke.

These methods generally leave admissible transitions implicit, relying on the model to select the next action from in-context information. The resulting failure modes are well documented, and include planning hallucination~\citep{zhu2025knowagent,xiao2024flowbench}, trajectory drift on long-horizon tasks, and repetitive loops when execution feedback is ambiguous. These failures motivate the explicit structural priors discussed next.

\subsection{Structured Priors for Agent Planning}
\label{app:priors-extended}

One approach is to represent procedures explicitly. KnowAgent~\citep{zhu2025knowagent} maintains a textual action knowledge base of admissible action rules and pairs it with knowledgeable self-learning to constrain the agent's action path during trajectory synthesis. FlowBench~\citep{xiao2024flowbench} formalizes workflow knowledge in three formats, text, code, and flowchart, and shows empirically across six domains and 51 scenarios that flowcharts reduce planning hallucination most effectively. AFlow~\citep{zhang2025aflow} recasts workflow construction as Monte Carlo Tree Search over code-represented graphs whose nodes are LLM-invoking operators. Decoding constraints provide another form of structure: TOOLDEC~\citep{zhang2023don} compiles tool syntax schemas into finite-state automata and constrains generation to syntactically valid calls, using the efficient FSM-guided generation algorithm of Outlines~\citep{willard2023efficient}. Such methods guarantee surface-form validity but say nothing about whether an action is semantically admissible given the task state.

Other methods organize tool collections as graphs. ToolNet~\citep{liu2024toolnet} mines a directed tool-transition graph from LLM-generated trajectories and lets the agent walk it at inference time. ControlLLM~\citep{liu2024controlllm} pre-builds a tool dependency graph from parameter-type matching and introduces \emph{Thoughts-on-Graph} search over it. COLT~\citep{qu2024towards} targets retrieval completeness through a dual-view query-tool-scene graph trained with LightGCN and contrastive losses. Graph RAG-Tool Fusion~\citep{lumer2025graph} hybridizes vector retrieval with graph traversal over a hand-designed tool knowledge graph, extending GraphRAG~\citep{edge2024local} to tool selection. The Tool Graph Retriever~\citep{gao2025tool} learns a tool-dependency discriminator and propagates embeddings over the resulting graph, while NaviAgent~\citep{jiang2025naviagent} fuses API schema structure with historical invocations into a continuously evolving heterogeneous dependency graph. On the evaluation side, TaskBench~\citep{shen2024taskbench} provides a graph-structured tool-automation benchmark with explicit node and edge scoring, and GNN4TaskPlan~\citep{wu2024can} demonstrates that GNN-based sub-task selection improves over LLM-only planning on it. SkillGraph~\citep{nie2026skillgraph} addresses multimodal multi-agent collaboration: it retrieves reasoning skills from an evolving skill bank and predicts a query-conditioned communication graph over agents.

These methods use graphs for different purposes: tool-transition and dependency graphs support tool selection, workflow graphs organize operator execution, and SkillGraph models communication among agents. Conditional action knowledge also appears in KnowAgent's textual rules and FlowBench's branching workflows. PG combines a graph over tool calls and reasoning steps with condition, guidance, and pitfall attributes on transitions. At each step, it localizes the current procedure and verbalizes the surrounding neighborhood. Per-step retrieval is also used by AutoGuide, which selects state-matched guidelines (Appendix~\ref{app:baseline_details}); PG instead retrieves connected transitions and supports explicit edits to their topology and attributes.

\subsection{Self-Improving Agents from Trajectories}
\label{app:trajectory-extended}

Another line learns reusable knowledge from the agent's execution history, often through verbal reflection and episodic memory. Reflexion~\citep{shinn2023reflexion} writes self-critiques into an episodic buffer and re-attempts the task; Generative Agents~\citep{park2023generative} maintain a memory stream ranked by recency, importance, and relevance, with periodic reflection for abstraction; MemoryBank~\citep{zhong2024memorybank} keeps a long-term store of experience summaries governed by an Ebbinghaus-inspired forgetting schedule; and RAP~\citep{kagaya2024rap} retrieves whole past trajectories as in-context exemplars. Two methods push toward explicitly contrastive distillation: ExpeL~\citep{zhao2024expel} contrasts success and failure pairs to extract natural-language insights, and AutoGuide~\citep{fu2024autoguide} sharpens these into \emph{context-aware guidelines} of explicit conditional form (``in context $X$, action $Y$ is appropriate'') retrieved at test time from the agent's current state. ERL~\citep{shi2026experiential} uses reflection to guide a second attempt, then trains the base policy to retain the resulting improvements.

A second family stores reusable skills and workflows. Voyager~\citep{wang2023voyager} maintains an ever-growing library of Minecraft skills indexed by embeddings of their natural-language descriptions, retrieved top-$k$ per task; TroVE~\citep{wang2024trove} induces a verified Python toolbox and trims it to stay compact; AWM~\citep{wang2025agent} induces reusable workflows combining natural-language descriptions with program-form actions and adds them to prompt memory in both offline and online modes; and SkillWeaver~\citep{zheng2025skillweaver} and WebXSkill~\citep{wang2026webxskill} refine skill discovery and execution for web agents. Several frameworks explicitly model the lifecycle of procedural memory: MemP~\citep{fang2025memp} formalizes build, retrieve, and update as an optimization target, distilling trajectories into both fine-grained step instructions and higher-level script abstractions; EvolveR~\citep{wu2025evolver} closes the loop with offline self-distillation, online retrieval of strategic principles, and policy reinforcement; and SEAgent~\citep{sun2025seagent} learns computer-use policies from autonomously collected experience. A-Mem~\citep{xu2026mem} and Zep/Graphiti~\citep{rasmussen2025zep} bring knowledge-graph-style structure to agent memory, though their focus is episodic and semantic rather than procedural.

Among these methods, AutoGuide~\citep{fu2024autoguide} is the closest to our approach. Like our edge attributes, its conditional guidelines associate situations with actions. PG connects these transitions in a typed graph, supporting structural retrieval, dependency inspection, and refinement through graph edits. MemP shares our emphasis on lifecycle operations over procedural memory, storing trajectories and script-like abstractions for retrieval without an explicit procedure graph. The seven baselines we compare empirically in Section~\ref{sec:experiments} are drawn from across this design space, spanning episodic summarization (MemoryBank), trajectory retrieval (RAP), insight distillation (ExpeL), conditional guidelines (AutoGuide), workflow induction (AWM), and textual transition rules (KnowAgent); Table~\ref{tab:baseline_summary} summarizes their storage and injection designs.

\subsection{Detailed Comparison Table}
\label{app:comparison}

Table~\ref{tab:method-comparison} compares 24 representative methods with PG along eight design dimensions. It summarizes the representations and access mechanisms described in the cited work; the baseline configurations used in our experiments are specified separately in Appendix~\ref{app:baseline_details}.

The comparison highlights differences in what is stored, how it is accessed, and what can be updated. Textual rules, code libraries, workflows, and graphs each preserve useful procedural structure. For example, KnowAgent supplies action-transition knowledge as text in the prompt, while FlowBench includes flowcharts with branch conditions. AFlow searches over code-represented operator workflows, and tool graphs support navigation through tool catalogs. PG stores transitions between tool calls and reasoning steps, with conditions, guidance, and pitfalls attached to edges, and retrieves a neighborhood around the current procedure.

The target of improvement also differs across methods. Some revise insights, skills, or workflows; ToolNet derives tool transitions from trajectories, while SkillGraph couples skill-bank updates with query-conditioned communication among agents. PG updates the procedure graph itself through node and edge edits. Its contribution is this combination of attributed transitions, localized guidance, and structural refinement, rather than graph structure or conditional knowledge alone.

\paragraph{Column Definitions.}
``Form'' describes the representation used for action selection, memory, or coordination; ``Granularity'' is the unit size of stored knowledge; and ``Source'' describes how the structure is obtained. ``Updatable'' indicates whether the structure can be revised from new trajectories, and ``Retrieval'' specifies the inference-time access mode. ``Edge Semantics'' describes the meaning of edges or relations, if any, while ``Scope'' identifies the intended deployment setting. ``Graph'' distinguishes explicit workflow, tool, or communication graphs (yes), auxiliary hierarchies or transitions encoded only in text (partial), and representations without explicit graph organization (no). FlowBench is shown in its flowchart form. For KnowAgent, ``Updatable'' refers to the action knowledge base, not model self-training. In our implementation, PG edge attributes comprise conditions, guidance, and pitfalls.

\begin{table}[htbp]
\centering
\scriptsize
\setlength{\aboverulesep}{0pt}
\setlength{\belowrulesep}{0pt}
\renewcommand{\arraystretch}{1.2} %
\setlength{\tabcolsep}{2.4pt}
\caption{\textbf{Comparison of representative methods along eight design dimensions.} Column definitions and method-specific conventions are described in the accompanying text.}
\label{tab:method-comparison}
\newcommand{\centeredstack}[1]{\ensuremath{\vcenter{\offinterlineskip\vskip1pt\hbox{\shortstack[l]{#1}}\vskip1pt}}}
\begin{tabular}{lllllllll}
\toprule
\rowcolor{GoogleGreen!15}
\textbf{Method} & \textbf{Form} & \textbf{Granularity} & \textbf{Source} & \textbf{Updatable} & \textbf{Retrieval} & \textbf{Edge Semantics} & \textbf{Scope} & \textbf{Graph} \\
\midrule
ReAct~\citep{yao2023react}             & none       & action     & none           & none      & none         & none              & general & no \\
Reflexion~\citep{shinn2023reflexion}   & text       & reflection & trajectory     & yes       & \centeredstack{prompt\\buffer} & none              & task    & no \\
MemoryBank~\citep{zhong2024memorybank} & text       & summary    & trajectory     & yes       & embedding    & none              & task    & no \\
RAP~\citep{kagaya2024rap}              & text       & trajectory & trajectory     & yes       & embedding    & none              & task    & no \\
Toolformer~\citep{schick2023toolformer}& weights    & API call   & trajectory     & no        & none         & none              & catalog & no \\
CodeAct~\citep{wang2024executable}          & code       & action     & manual         & no        & none         & none              & general & no \\
ToolLLM~\citep{qin2024toolllm}         & document   & API        & manual         & no        & embedding    & none              & catalog & no \\
Gorilla~\citep{patil2024gorilla}       & document   & API        & manual         & no        & embedding    & none              & catalog & no \\
AnyTool~\citep{du2024anytool}          & tree       & API        & manual         & no        & structural   & taxonomy          & catalog & partial \\
ToolGen~\citep{wang2025toolgen}        & tokens     & API        & trajectory     & no        & none         & none              & catalog & no \\
Voyager~\citep{wang2023voyager}        & code       & skill      & trajectory     & yes       & embedding    & none              & task    & no \\
TroVE~\citep{wang2024trove}            & code       & skill      & trajectory     & yes       & \centeredstack{prompt /\\import} & none              & task    & no \\
AWM~\citep{wang2025agent} & \centeredstack{text +\\code} & workflow & trajectory & yes & \centeredstack{prompt\\memory} & sequence & task & no \\
ExpeL~\citep{zhao2024expel}            & text       & insight    & trajectory     & yes       & embedding    & none              & task    & no \\
AutoGuide~\citep{fu2024autoguide}      & text       & rule       & trajectory     & yes       & \centeredstack{LLM\\selection} & implicit cond.    & task    & no \\
MemP~\citep{fang2025memp}              & \centeredstack{text +\\trajectories} & procedure  & trajectory     & yes       & embedding    & none              & task    & no \\
EvolveR~\citep{wu2025evolver}          & text       & principle  & trajectory     & yes       & embedding    & none              & task    & no \\
KnowAgent~\citep{zhu2025knowagent} & text & action & hybrid & no & \centeredstack{static\\prompt} & action transition & task & partial \\
FlowBench~\citep{xiao2024flowbench} & flowchart & workflow & manual & no & prompt & branches & task & yes \\
AFlow~\citep{zhang2025aflow} & code & operator & trajectory & yes & execution & \centeredstack{control/\\data flow} & task & yes \\
ToolNet~\citep{liu2024toolnet}         & graph      & tool       & trajectory     & yes       & structural   & co-occurrence     & catalog & yes \\
ControlLLM~\citep{liu2024controlllm}   & graph      & tool       & manual         & no        & structural   & parameter dep.    & catalog & yes \\
Graph RAG-Tool~\citep{lumer2025graph}  & graph      & tool       & hybrid         & no        & hybrid       & dependency        & catalog & yes \\
SkillGraph~\citep{nie2026skillgraph} & \centeredstack{graph +\\text} & agent/skill & hybrid & yes & embedding & communication & \centeredstack{multi-\\agent} & yes \\
\midrule
\rowcolor{GoogleGreen}
\textcolor{white}{\textbf{Procedural Graph (ours)}} & \textcolor{white}{\textbf{graph}} & \textcolor{white}{\textbf{action}} & \textcolor{white}{\textbf{hybrid}} & \textcolor{white}{\textbf{yes}} & \textcolor{white}{\textbf{hybrid}} & \textcolor{white}{\centeredstack{\textbf{transition}\\\textbf{attributes}}} & \textcolor{white}{\textbf{task}} & \textcolor{white}{\textbf{yes}} \\
\bottomrule
\end{tabular}
\end{table}

\section{Experimental Details}
\label{app:experiment_details}

This appendix provides the dataset splits, evaluation metrics, baseline implementation details, prompt templates, and the full self-evolution algorithm (Appendix~\ref{app:evolution_algorithm}).

\subsection{Datasets and Splits}
\label{app:datasets}

Table~\ref{tab:dataset_splits} summarizes the sample counts and splits used in all experiments.

\begin{table}[htbp]
\centering
\scriptsize
\setlength{\aboverulesep}{0pt}
\setlength{\belowrulesep}{0pt}
\renewcommand{\arraystretch}{1.2}
\setlength{\tabcolsep}{2pt}
\caption{\textbf{Dataset statistics and splits.}}
\label{tab:dataset_splits}
\newcommand{\dset}[2]{#1 \newline \citep{#2}}
\begin{tabular}{>{\raggedright\arraybackslash}m{3.3cm} >{\raggedright\arraybackslash}m{2.5cm} r r >{\raggedright\arraybackslash}m{8.7cm}}
\toprule
\rowcolor{GoogleGreen!15}
\textbf{Dataset} & \textbf{Domain} & \textbf{Train} & \textbf{Test} & \textbf{Notes} \\
\midrule
\dset{HotpotQA}{yang2018hotpotqa} & Open-domain QA & 1,000 & 1,000 & Disjoint samples from the official validation pool, de-duplicated by question id. \\
\dset{MultiChallenge}{deshpande2025multichallenge} & Multi-turn dialogue & 100 & 166 & The main table evaluates all 166 test items; the construction study (Table~\ref{tab:performance_metrics}) uses the 56-sample fast split. \\
\dset{GDPval}{patwardhan2025gdpval} & Professional deliverables & 88 & 44 & All tasks split deterministically by occupation. \\
\dset{ALFWorld}{shridharalfworld} & Embodied household & 238 & 134 & Test uses the standard \emph{unseen} split; loaded counts reflect internal filtering by the ALFWorld library. \\
\dset{$\tau$-bench}{yao2024tau} & Customer service & 500 & 115 & Test uses the \emph{retail} domain. \\
\dset{BFCL v3}{patil2025berkeley} & Function calling & 100 & 100 & Base-category multi-turn tasks. \\
\dset{EnterpriseArena}{han2026can} & Financial simulation & 50 & 50 & Simulator episodes under disjoint random-seed sets. The self-evolution study (Table~\ref{tab:evolution_rounds}) uses a smaller $20/20/20$ configuration. \\
\bottomrule
\end{tabular}
\end{table}

During self-evolution, the training split is processed in sequential strides of $S{=}100$ samples on HotpotQA and $S{=}20$ on MultiChallenge. The validation set $\mathcal{D}_{\text{val}}$ consumed by the acceptance gate is held out separately from both splits above and never overlaps the test set; its size is $1{,}000$ for HotpotQA and $100$ for MultiChallenge (Appendix~\ref{app:construction_setup}), and $20$ episodes for EnterpriseArena. For EnterpriseArena, each configuration runs $50$ training and $50$ test episodes; the full environment mechanics are given in Appendix~\ref{app:cfo_rules}.

\subsection{Evaluation Metrics}
\label{app:metric_details}

For \textbf{HotpotQA}, the main comparison reports LLM-judged answer accuracy: a Gemini 3.1 Pro judge model receives the question, the gold answer, and the agent's answer, and returns a binary equivalence verdict; the construction study (Section~\ref{sec:pg_construction}) additionally reports strict string Exact Match and word-level F1. For \textbf{MultiChallenge}, an LLM judge based on Gemini 3.1 Pro scores the Overall Success Rate together with four axes: Inference Memory, Instruction Retention, Reliable Versioned Editing, and Self Coherence. For \textbf{GDPval}, deliverables are scored against per-task rubrics and we report the mean rubric score. For \textbf{ALFWorld}, we report the task success rate on the test games. For \textbf{$\tau$-bench}, we report Pass@1, i.e., the fraction of episodes whose final database state matches the annotated goal state. For \textbf{BFCL v3}, we report the official multi-turn accuracy. For \textbf{EnterpriseArena}, we report the full-horizon survival rate, the average lifespan in months, the mean time-averaged enterprise score, and the average capital raised (see Appendix~\ref{app:cfo_rules} for definitions).

\subsection{Baseline Implementation Details}
\label{app:baseline_details}

All baselines share the same ReAct solver, tool interface, and decoding configuration as our method, and every learning-based baseline consumes exactly the same training split that our self-evolution loop uses; they differ only in the artifact distilled from those trajectories and in how that artifact is injected at inference time. Table~\ref{tab:baseline_summary} summarizes the compared mechanisms: what each method stores, how that knowledge is organized, and how it enters the solver's context at inference time.

\begin{table}[htbp]
\centering
\scriptsize
\setlength{\aboverulesep}{0pt}
\setlength{\belowrulesep}{0pt}
\renewcommand{\arraystretch}{1.2}
\setlength{\tabcolsep}{3pt}
\caption{\textbf{Memory mechanisms compared in the main experiments.}}
\label{tab:baseline_summary}
\begin{tabular}{l >{\raggedright\arraybackslash}m{3.7cm} >{\raggedright\arraybackslash}m{4cm} >{\raggedright\arraybackslash}m{3.8cm}}
\toprule
\rowcolor{GoogleGreen!15}
\textbf{Method} & \textbf{Stored Artifact} & \textbf{Organization} & \textbf{Used at Inference} \\
\midrule
Vanilla ReAct~\citep{yao2023react} & none & -- & -- \\
MemoryBank~\citep{zhong2024memorybank} & experience summaries & unstructured pool with forgetting & retrieved and prepended \\
RAP~\citep{kagaya2024rap} & raw trajectories & similarity index & top-$k$ in-context exemplars \\
ExpeL~\citep{zhao2024expel} & natural-language insights & unordered list & injected into system prompt \\
AutoGuide~\citep{fu2024autoguide} & ``in state $X$, do $Y$'' guidelines & condition-indexed pool & state-matched retrieval \\
AWM~\citep{wang2025agent} & workflows with text and actions & sequence library & workflow memory in prompt \\
KnowAgent~\citep{zhu2025knowagent} & action rules and transitions & single text document & static prompt prefix \\
\rowcolor{GoogleGreen}
\textcolor{white}{\textbf{Procedural Graph (Ours)}} & \textcolor{white}{\textbf{(procedure, relation, procedure) triplets}} & \textcolor{white}{\textbf{connected graph}} & \textcolor{white}{\textbf{localized subgraph $\rightarrow$ generated guidance}} \\
\bottomrule
\end{tabular}
\end{table}

\textbf{MemoryBank}~\citep{zhong2024memorybank} maintains a long-term store of per-task experience summaries updated after each completed task; at inference, relevant summaries are retrieved with recency-weighted relevance and prepended to the solver prompt.

\textbf{RAP}~\citep{kagaya2024rap} embeds completed trajectories and, at each new task, retrieves the most similar past trajectories as in-context exemplars.

\textbf{ExpeL}~\citep{zhao2024expel} contrasts success and failure trajectories to distill a pool of natural-language insights, which are injected into the system prompt alongside retrieved successful exemplars.

\textbf{AutoGuide}~\citep{fu2024autoguide} extracts state-conditioned guidelines from contrastive trajectory pairs; at each step, the current state is summarized and the applicable guidelines are retrieved and injected.

\textbf{AWM}~\citep{wang2025agent} induces reusable workflows from successful trajectories, retaining natural-language descriptions and action sequences as workflow memory in the solver prompt.

\textbf{KnowAgent}~\citep{zhu2025knowagent} maintains a textual action-knowledge base describing available actions and admissible transition rules, injected as a static prompt prefix.

\subsection{Procedural Graph Statistics}
\label{app:pg_stats}

Table~\ref{tab:pg_stats} summarizes the size of the Procedural Graph used for each benchmark in the main experiments. The graphs are compact: outside of BFCL~v3, whose $131$ nodes mirror its large function catalog, every graph has between $7$ and $17$ nodes and between $7$ and $27$ triplets. Across all graphs, the relation vocabulary $\mathcal{R}$ comprises four types: \texttt{LEADS\_TO}, \texttt{TRIGGERS}, \texttt{PROVIDES\_INPUT\_FOR}, and \texttt{CONVERGES\_TO}. Most edges carry the full \emph{condition}/\emph{guidance}/\emph{pitfalls} attribute triple of Section~\ref{subsec:graph_representation}, with \emph{guidance} the most consistently populated field.

\begin{table}[htbp]
\centering
\footnotesize
\setlength{\aboverulesep}{0pt}
\setlength{\belowrulesep}{0pt}
\renewcommand{\arraystretch}{1.2}
\setlength{\tabcolsep}{8pt}
\caption{\textbf{Sizes of the Procedural Graphs used in the main experiments.}}
\label{tab:pg_stats}
\begin{tabular}{l ccccccc}
\toprule
\rowcolor{GoogleGreen!15}
 & \textbf{HotpotQA} & \textbf{MultiChallenge} & \textbf{GDPval} & \textbf{ALFWorld} & \textbf{$\tau$-bench} & \textbf{BFCL v3} & \textbf{EnterpriseArena} \\
\midrule
\textbf{Nodes}    & 9 & 7 & 15 & 11 & 17 & 131 & 11 \\
\textbf{Triplets} & 9 & 7 & 22 & 27 & 18 & 265 & 13 \\
\bottomrule
\end{tabular}
\end{table}

\subsection{Prompt Templates}
\label{app:prompts}

We use three families of prompts: a solver execution prompt that governs the ReAct loop, guidance generation prompts that translate the (sub)graph into situational guidance at each step, and a refiner prompt that drives offline self-evolution. The same templates are shared across all benchmarks; only the tool lists and task descriptions vary. Curly braces denote runtime placeholders.

\newcommand{\pvar}[1]{\textcolor{GoogleGreen!50!black}{\texttt{\{#1\}}}}

\begin{tracebox}{GoogleGreen}{Solver Execution Prompt (ReAct loop)}
\footnotesize\raggedright
\pvar{system\_prompt}

\smallskip
\textbf{Procedural Graph Guidance:} \pvar{procedural\_graph\_guidance}

\smallskip
You must interleave Thought and Action. Your output format must be exactly:\\
\texttt{Thought: <your reasoning about what to do next>}\\
\texttt{Action: <tool\_name>(arg1=val1, arg2=val2, ...)}

\smallskip
Example:\\
\texttt{Thought: I need to check the files in the workspace directory to locate the source documents.}\\
\texttt{Action: list\_dir(path=".")}

\smallskip
DO NOT write any ``Observation:'' block or any subsequent steps. Only output exactly one Thought and one Action block. Do NOT simulate the environment's responses.

\smallskip
\textbf{Current Trajectory:} \pvar{trajectory}

\smallskip
\texttt{Thought:}
\end{tracebox}

\begin{tracebox}{GoogleGreen}{Guidance Generation Prompt (Local subgraph; default)}
\footnotesize\raggedright

\medskip
You are an expert cognitive architect and execution guide for an AI agent solving the task: \pvar{task\_description}

\smallskip
\textbf{Here is} \pvar{graph\_context\_desc}\textbf{:} \pvar{subgraph\_summary}\\
\textbf{Here is the current active query / observation:} \pvar{query}\\
\textbf{Here is the agent's recent execution trajectory:} \pvar{recent\_context}

\medskip
Analyze this \pvar{graph\_source} in the context of the agent's current progress. Using the \texttt{condition}, \texttt{guidance}, and \texttt{pitfalls} attributes carried by the edges in the graph context, generate clear, detailed, and actionable guidance advising the agent on exactly what step or strategy to pursue next, what pitfalls to avoid, and how to recover from recent failures if any. You must include any specific command patterns, file paths, tools, or arguments defined in the graph context if they are relevant to the next steps.

\end{tracebox}

\paragraph{Serialized Graph Context Example.}
We reconstruct the graph-context text below from the saved HotpotQA Mode 2 graph using the implemented local serializer. The active-node header supplies the node ID, type, and description. Directed transitions are grouped by hop and followed by their condition, guidance, and pitfalls, preserving the checkpoint's field text. The two stored relation labels, \texttt{LEADS\_TO} and \texttt{PROVIDES\_INPUT\_FOR}, are not printed by this serializer.

\begin{tcolorbox}[colback=GoogleGreen!3,colframe=GoogleGreen,colbacktitle=GoogleGreen,coltitle=white,fonttitle=\bfseries,title={Serialized Local Graph Context (HotpotQA; excerpt)},arc=3pt,outer arc=3pt,boxrule=0.8pt,left=8pt,right=8pt,top=6pt,bottom=6pt,before skip=8pt,after skip=8pt,breakable=false]
\footnotesize\raggedright\ttfamily
Active Cognitive Node: [First\_Hop\_Retrieve] (Type: ACTION)\par
Description: Execute first\_hop\_retrieve to fetch primary evidence passages.\par
\smallskip
Immediate Transition Options (Hop 1):\par
- Transition: [First\_Hop\_Retrieve] $\rightarrow$ [Scan\_Index] (Condition: first\_hop\_retrieve)\par
* Guidance: Review the retrieved primary passages via Scan\_Index to locate specific bridge terms (such as birth dates, locations, or associated entities).\par
* Pitfalls to Avoid: Do not skip reading evidence details; missing the exact bridge entity name causes second-hop search failure.\par
\smallskip
Subsequent Horizon (Hop 2):\par
- Transition: [Scan\_Index] $\rightarrow$ [Bridge\_Extract] (Condition: scan\_index)\par
* Guidance: Extract the explicit connecting entity or bridge term linking the first passage to the target question.\par
* Pitfalls to Avoid: Ensure the extracted bridge term matches exact Wikipedia capitalization conventions.\par
\end{tcolorbox}

\begin{tracebox}{GoogleGreen}{Guidance Generation Prompt (Full graph variant)}
\footnotesize\raggedright
The full-graph variant is \emph{textually identical} to the prompt above; the only difference is the content bound to the graph context slot, which is the complete Procedural Graph rather than the localized subgraph. Concretely, \pvar{graph\_context\_desc} is instantiated as ``the complete Procedural Graph governing the task structure and strategic guidance'', \pvar{subgraph\_summary} as the full \pvar{graph\_summary}, and \pvar{graph\_source} as ``complete Procedural Graph''. The task description, query, recent-trajectory window, analysis instruction, and output requirements are unchanged, so the two ablation rows in Table~\ref{tab:ablation_graph_configs} differ only in graph scope and not in prompt wording or requested output length.
\end{tracebox}

\begin{tracebox}{GoogleGreen}{Refiner Prompt (Self-evolution)}
\footnotesize\raggedright
You are an expert cognitive architect optimizing a Procedural Graph for an intelligent agent. The Procedural Graph encodes structured procedural guidance.

\smallskip
\textbf{Task context:} \pvar{task\_description}\\
\textbf{Refinement mode:} \pvar{mode}\\
\textbf{Available Tool Actions} (the agent can only execute these actions): \pvar{available\_tools\_list}\\
\textbf{Recent execution trajectories:} \pvar{attempts\_block}\\
\textbf{Current Procedural Graph representation:} \pvar{current\_graph\_json}\\
\textbf{Previously rejected candidates:} \pvar{rejected\_block}

\medskip
Your job is to refine the Procedural Graph. Follow these guidelines based on the mode:
\begin{itemize}[leftmargin=12pt, itemsep=1pt, topsep=2pt]
    \item \texttt{static\_onetime / static\_incremental}: Prune edges/nodes that lead to loops, deadlocks, or failures. Add missing nodes and edges that could fix the failures and improve performance for future tasks.
    \item \texttt{scratch\_onetime / scratch\_incremental}: If starting from scratch (the graph contains only $\text{Start} \rightarrow \text{End}$), synthesize a brand new, complete Procedural Graph using the Available Tool Actions list, Status, and successful patterns in the trajectories. Otherwise, prune edges/nodes that lead to loops, deadlocks, or failures, and add missing nodes and edges based on the given graph.
\end{itemize}

\textbf{Rules for nodes and edges.} Rules 2--4 describe the edge attributes in $\Phi(e)$: \texttt{condition}, \texttt{guidance}, and \texttt{pitfalls}. The remaining rules govern node compatibility, generality, and graph structure.
\begin{enumerate}[leftmargin=14pt, itemsep=1pt, topsep=2pt]
    \item \textbf{Action Nodes.} Any node of type \texttt{ACTION} must match one of the action/tool names in the ``Available Tool Actions'' list above.
    \item \textbf{Transition Conditions.} If an edge has a \texttt{condition}, provide a natural-language semantic precondition under which this transition should fire (e.g., ``When dialogue history has been parsed but target constraints are unknown''). Use \texttt{null} if the transition is unconditional.
    \item \textbf{Execution Guidance.} For every edge added in \texttt{add\_edges}, you MUST provide a \texttt{guidance} string detailing exactly what action to take next and the strategic rationale behind it.
    \item \textbf{Pitfalls.} Provide a \texttt{pitfalls} string warning about premature actions, forbidden words, or common formatting pitfalls to avoid during this step.
    \item \textbf{Generality \& Leak Prevention.} The updated Procedural Graph must guide the agent effectively without overfitting to specific details of a single trajectory. Use high-level conceptual descriptions. 
    \item \textbf{Node ID Compatibility.} If refining an existing graph (static modes), you MUST preserve the existing node IDs (such as \texttt{Month\_Start}, \texttt{Decide\_Capital}, and the tool names) so they remain compatible with the environment's state tracker. Do not rename them.
    \item \textbf{Graph Structure.} Follow the task's configured cycle policy. Every edge must reference existing nodes, and every node must have a directed path to a terminal node. The environment loop handles repetition across simulation cycles.
\end{enumerate}

Please propose the exact set of edits to perform. You must output your edits as a single valid JSON block containing four arrays: \texttt{add\_nodes}, \texttt{delete\_nodes}, \texttt{add\_edges}, and \texttt{delete\_edges}. Output format must be exactly:

\smallskip
\texttt{\{ "add\_nodes": \hspace{2pt}[\{"id":..., "type": "ACTION", "description":...\}],}\\
\texttt{\hspace*{9pt}"delete\_nodes": ["node\_id"],}\\
\texttt{\hspace*{9pt}"add\_edges": \hspace{2pt}[\{"source":..., "target":..., "relation":...,}\\
\texttt{\hspace*{86pt}"condition":..., "guidance":..., "pitfalls":...\}],}\\
\texttt{\hspace*{9pt}"delete\_edges": [\{"source":..., "target":...\}] \}}

\smallskip
Make sure to output ONLY the raw JSON block.
\end{tracebox}

\begin{samepage}
Each entry in \texttt{delete\_edges} removes all edges with the specified source and target, regardless of relation. To retain selected transitions between the same endpoints, include them in \texttt{add\_edges}, which is applied after deletion.
\end{samepage}

\subsection{Self-Evolution Algorithm}
\label{app:evolution_algorithm}

Algorithm~\ref{alg:evolution} makes the retained-checkpoint state explicit. $S_k$ is the cached validation score of $\mathcal{G}_k$, and $c$ specifies whether cycles are allowed. $\mathrm{Tail}_{L_{\max}}$ preserves the ending of its input by removing excess tokens from the beginning, leaving shorter inputs unchanged. The graph stays fixed during each training or validation episode. $\mathrm{SerializeRejections}$ supplies prior candidate graphs and their validation scores, or structural-failure diagnostics, to the refiner; the associated training traces remain part of the rejection record.

\begingroup
\setlength{\intextsep}{5pt}
\begin{algorithm}[H]
\small
\caption{Offline Closed-Loop Procedural Graph Self-Evolution}
\label{alg:evolution}
\begin{algorithmic}[1]
    \REQUIRE Initial graph $\mathcal{G}_0$; training/validation sets $\mathcal{D}_{\mathrm{train}},\mathcal{D}_{\mathrm{val}}$;
    round budget $K$; trajectory limit $L_{\max}$; cycle policy $c$.
    \STATE $S_0 \leftarrow \mathrm{Evaluate}(\mathcal{G}_0,\mathcal{D}_{\mathrm{val}})$
    \STATE $\mathcal{H}_{\mathrm{rejected}} \leftarrow [\,]$
    \FOR{$k = 1,\ldots,K$}
        \STATE $\mathcal{G}_k \leftarrow \mathcal{G}_{k-1}$; $S_k \leftarrow S_{k-1}$ \COMMENT{Retain unless accepted}
        \STATE Select training batch $\mathcal{B}_k \subset \mathcal{D}_{\mathrm{train}}$
        \STATE $\mathcal{E}_k \leftarrow \mathrm{Rollout}(\mathcal{G}_{k-1},\mathcal{B}_k)$ \COMMENT{Training traces and scores}
        \STATE $\mathcal{C}_k \leftarrow \mathrm{Tail}_{L_{\max}}(\mathrm{ConcatTrajectories}(\mathcal{E}_k))$
        \STATE $\mathcal{R}_k \leftarrow \mathrm{SerializeRejections}(\mathcal{H}_{\mathrm{rejected}})$
        \STATE $\Delta\mathcal{G}_k \leftarrow \mathrm{Refiner}(\mathcal{G}_{k-1},\mathcal{C}_k,\{S_i^{(k)}\}_i,\mathcal{R}_k)$
        \STATE $(\mathcal{G}_k^{\mathrm{cand}},d_k) \leftarrow \mathrm{PrepareCandidate}(\mathcal{G}_{k-1},\Delta\mathcal{G}_k,c)$
        \IF{$d_k \neq \varnothing$}
            \STATE Append $(\Delta\mathcal{G}_k,\mathcal{G}_k^{\mathrm{cand}},\mathcal{E}_k,d_k)$ to $\mathcal{H}_{\mathrm{rejected}}$
            \STATE \textbf{continue} \COMMENT{No validation rollout; retained state is unchanged}
        \ENDIF
        \STATE $S_k^{\mathrm{cand}} \leftarrow \mathrm{Evaluate}(\mathcal{G}_k^{\mathrm{cand}},\mathcal{D}_{\mathrm{val}})$
        \IF{$S_k^{\mathrm{cand}} \ge S_{k-1}$}
            \STATE $\mathcal{G}_k \leftarrow \mathcal{G}_k^{\mathrm{cand}}$; $S_k \leftarrow S_k^{\mathrm{cand}}$ \COMMENT{Accept, including ties}
        \ELSE
            \STATE Append $(\Delta\mathcal{G}_k,\mathcal{G}_k^{\mathrm{cand}},\mathcal{E}_k,S_k^{\mathrm{cand}})$ to $\mathcal{H}_{\mathrm{rejected}}$
        \ENDIF
    \ENDFOR
    \RETURN $\mathcal{G}_K$
\end{algorithmic}
\end{algorithm}
\endgroup

\paragraph{Candidate Preparation and Structural Checks.}
$\mathrm{PrepareCandidate}$ applies edits to a copy of the retained graph, deleting edges and nodes before adding nodes and edges. It reports malformed edits, invalid node or relation types, and missing edge endpoints as failures. When cycles are disallowed, the implementation removes detected cycle-closing edges before validation; when cycles are allowed, that repair and the acyclicity check are skipped. The remaining checks require valid edge endpoints and a directed path from every node to a terminal node, defined by zero out-degree. This is a reachability check to a terminal node, not specifically to the node named \texttt{End}. Matching action-node names to the available tool list is a refiner-prompt requirement; the generic structural validator does not independently enforce tool-catalog membership. On failure, $d_k$ contains diagnostics and $\mathcal{G}_k^{\mathrm{cand}}$ may be unavailable; on success, $d_k=\varnothing$.

\section{Long-Horizon Analysis on EnterpriseArena}
\label{app:cfo_appendix}

This appendix describes the EnterpriseArena simulator and evaluation metrics, reports survival and cash trajectories, and compares traces from the baseline, memory-summarization, and PG-guided agents.

\subsection{EnterpriseArena Mechanics and Crisis Schedule}
\label{app:cfo_rules}

The Chief Financial Officer (CFO) simulator models the balance sheet dynamics of a microfinance lending institution over a long-term horizon of up to $132$ months. The state of the environment at month $t$ is formalized as a multi-dimensional tuple:
\begin{equation}
    \mathcal{S}_t = \big( C_t, \, L_t, \, A_t, \, IR_t, \, PR_t, \, AP_t, \, D_t, \, E_t, \, U_t \big),
\end{equation}
where $C_t$ is the cash balance, $L_t$ is the gross loan portfolio, $A_t$ is the allowance for loan losses, $IR_t$ and $PR_t$ are interest and principal receivables, $AP_t$ is accounts payable, $D_t$ is total outstanding debt, $E_t$ is total equity raised, and $U_t$ is the active user base.

The agent interacts with the simulator through a discrete action space. The primary state-advancing action is \texttt{book\_closing()}, which simulates the transition from month $t$ to $t+1$. During this transition, the environment executes the following operations:
\begin{enumerate}[leftmargin=*]
    \item \textbf{Loan Amortization}: A fraction of the loan portfolio $L_t$ matures, generating principal payments and interest income based on the lending rate.
    \item \textbf{User and Operational Costs}: The active user base $U_t$ grows or decays organically. Fixed operational costs and user acquisition costs are deducted from the cash balance $C_t$.
    \item \textbf{Write-Offs}: Defaulted loans are written off against the allowance $A_t$, and new provisions are calculated.
\end{enumerate}

To manage liquidity, the agent can invoke \texttt{fund\_raising\_request(type, amount)}, where the type is either \texttt{'equity'} or \texttt{'debt'}. Fundraising is subject to two realistic constraints:
\begin{itemize}[leftmargin=*]
    \item \textbf{Market Delivery Lag}: Capital is not delivered immediately. There is a stochastic delay of $1$ to $6$ months between the request step and the cash injection.
    \item \textbf{Market Capacity Cap}: The maximum amount of capital that can be raised in a single request is dynamically capped by the environment based on the current macroeconomic state and the institution's financial health.
\end{itemize}

The simulation terminates immediately if the cash balance goes negative ($C_t < 0$), representing corporate bankruptcy. The environment simulates three successive macroeconomic crises to test the agent's long-term resilience:
\begin{itemize}[leftmargin=*]
    \item \textbf{Crisis 1 (Month 32)}: A mild contraction where loan repayment rates drop slightly from $98\%$ to $90\%$.
    \item \textbf{Crisis 2 (Month 59)}: A severe economic recession. Repayment rates plunge to $60\%$, write-offs surge, and organic user growth turns negative.
    \item \textbf{Crisis 3 (Month 112)}: A systemic liquidity freeze. Repayment rates drop to $40\%$, and the market capacity cap for fundraising is severely restricted, making new capital acquisition extremely difficult.
\end{itemize}

\paragraph{Evaluation Metrics.}
\textbf{Full Surv.} is the percentage of runs that complete the $132$-month horizon without bankruptcy; the crisis columns report the fractions reaching months $32$, $59$, and $112$. \textbf{Avg. Months} averages run duration, including early terminations. \textbf{Tools/Mo} averages the per-run ratio of arena information-tool calls to simulated months, excluding memory operations and state-changing actions. \textbf{Raised} is the mean cumulative equity and debt financing actually received per run, in millions of dollars.

We use EnterpriseArena's revenue-based valuation and tool-use penalty~\citep{han2026can} to compute the score at each recorded month:
\begin{equation}
    s_{i,t} =
    \begin{cases}
        0, & \text{if run } i \text{ has gone bankrupt}, \\
        5\,\mathrm{Rev}^{(12)}_{i,t} + C_{i,t} - 5{,}000\,N_{i,t}, & \text{otherwise},
    \end{cases}
    \label{eq:enterprise_score}
\end{equation}
where $\mathrm{Rev}^{(12)}_{i,t}$ is trailing-twelve-month revenue (annualized from the available monthly average when fewer than twelve months are recorded), $C_{i,t}$ is cash in dollars, and $N_{i,t}$ is the cumulative number of arena information-tool calls. \textbf{Avg. Score} first averages these monthly scores within each run up to termination, then averages across runs, reporting the result in millions of dollars. Thus, a run ending in bankruptcy can still have a positive time-averaged score.

\subsection{Results and Survival Analysis}
\label{app:cfo_plots}

\begin{table*}[htbp]
\centering
\scriptsize
\setlength{\aboverulesep}{0pt}
\setlength{\belowrulesep}{0pt}
\renewcommand{\arraystretch}{1.2}
\setlength{\tabcolsep}{2.55pt}
\caption{\textbf{Performance comparison on EnterpriseArena.} Within each model block, higher survival, lifespan, and score and lower Tools/Mo are highlighted. Raised highlights the largest financing volume. Metric definitions are given in Appendix~\ref{app:cfo_rules}.}
\label{tab:gemini_grok_simulation}
\begin{tabular}{l c c c c c c c c}
\toprule
\rowcolor{GoogleRed!15}
\cellcolor{GoogleRed!15} & \multicolumn{3}{c}{\textbf{Overall}} & \multicolumn{3}{c}{\textbf{Multi-Crisis Survival}} & \multicolumn{2}{c}{\textbf{Agent Performance}} \\

\arrayrulecolor{white}
\cmidrule(lr){2-4} \cmidrule(lr){5-7} \cmidrule(lr){8-9}
\arrayrulecolor{black}

\rowcolor{GoogleRed!15}
\multirow{-2}{*}{\textbf{Model \& Config}} & \textbf{Full Surv.} & \textbf{Avg. Months} & \textbf{Avg. Score} & \textbf{1st Crisis} & \textbf{2nd Crisis} & \textbf{3rd Crisis} & \textbf{Tools/Mo} & \textbf{Raised} \\
\midrule

\multicolumn{9}{l}{\textbf{Claude Sonnet 4.6}} \\
\midrule
\quad Baseline                      & 44.0\% & 89.80 & \cellcolor{GoogleRed}\textcolor{white}{\textbf{\$78.86M}} & \cellcolor{GoogleRed}\textcolor{white}{\textbf{100.0\%}} & 78.0\% & 52.0\% & 0.13 & \$152.12M \\
\quad RAP~\citep{kagaya2024rap}          & 50.0\% & 93.24 & \$76.76M & \cellcolor{GoogleRed}\textcolor{white}{\textbf{100.0\%}} & 70.0\% & 54.0\% & \cellcolor{GoogleRed}\textcolor{white}{\textbf{0.12}} & \cellcolor{GoogleRed}\textcolor{white}{\textbf{\$157.54M}} \\
\quad MemoryBank~\citep{zhong2024memorybank}              & 56.0\% & 97.82 & \$56.57M & \cellcolor{GoogleRed}\textcolor{white}{\textbf{100.0\%}} & 78.0\% & \cellcolor{GoogleRed}\textcolor{white}{\textbf{62.0\%}} & 0.27 & \$97.01M \\
\quad Procedural Graph (Ours)           & \cellcolor{GoogleRed}\textcolor{white}{\textbf{58.0\%}} & \cellcolor{GoogleRed}\textcolor{white}{\textbf{98.58}} & \$70.38M & \cellcolor{GoogleRed}\textcolor{white}{\textbf{100.0\%}} & \cellcolor{GoogleRed}\textcolor{white}{\textbf{80.0\%}} & 60.0\% & 0.36 & \$130.44M \\

\midrule

\multicolumn{9}{l}{\textbf{Gemini 3.1 Pro}} \\
\midrule
\quad Baseline                      & 6.0\% & 50.28 & \$31.85M & \cellcolor{GoogleRed}\textcolor{white}{\textbf{100.0\%}} & 38.0\% & 8.0\% & 0.89 & \$21.82M \\
\quad RAP~\citep{kagaya2024rap}             & 6.0\% & 51.34 & \$32.27M & \cellcolor{GoogleRed}\textcolor{white}{\textbf{100.0\%}} & 38.0\% & 12.0\% & 0.83 & \$23.16M \\
\quad MemoryBank~\citep{zhong2024memorybank}          & 22.0\% & 59.78 & \$33.11M & \cellcolor{GoogleRed}\textcolor{white}{\textbf{100.0\%}} & 38.0\% & 24.0\% & \cellcolor{GoogleRed}\textcolor{white}{\textbf{0.39}} & \$14.53M \\
\quad Procedural Graph (Ours)           & \cellcolor{GoogleRed}\textcolor{white}{\textbf{34.0\%}} & \cellcolor{GoogleRed}\textcolor{white}{\textbf{79.22}} & \cellcolor{GoogleRed}\textcolor{white}{\textbf{\$37.21M}} & \cellcolor{GoogleRed}\textcolor{white}{\textbf{100.0\%}} & \cellcolor{GoogleRed}\textcolor{white}{\textbf{54.0\%}} & \cellcolor{GoogleRed}\textcolor{white}{\textbf{46.0\%}} & 3.18 & \cellcolor{GoogleRed}\textcolor{white}{\textbf{\$38.20M}} \\

\midrule

\multicolumn{9}{l}{\textbf{Gemini 3.5 Flash}} \\
\midrule
\quad Baseline                      & 0.0\% & 33.58 & \$28.59M & \cellcolor{GoogleRed}\textcolor{white}{\textbf{100.0\%}} & 0.0\%  & 0.0\%  & 18.94 & \$0.00M \\
\quad RAP~\citep{kagaya2024rap}              & 0.0\% & 34.04 & \$28.75M & \cellcolor{GoogleRed}\textcolor{white}{\textbf{100.0\%}} & 2.0\%  & 0.0\%  & 17.71 & \$0.69M \\
\quad MemoryBank~\citep{zhong2024memorybank}          & 0.0\% & 35.60 & \cellcolor{GoogleRed}\textcolor{white}{\textbf{\$29.34M}} & \cellcolor{GoogleRed}\textcolor{white}{\textbf{100.0\%}} & 10.0\% & 0.0\%  & \cellcolor{GoogleRed}\textcolor{white}{\textbf{11.99}} & \$1.69M \\
\quad Procedural Graph (Ours)           & 0.0\% & \cellcolor{GoogleRed}\textcolor{white}{\textbf{40.62}} & \$29.08M & \cellcolor{GoogleRed}\textcolor{white}{\textbf{100.0\%}} & \cellcolor{GoogleRed}\textcolor{white}{\textbf{14.0\%}} & 0.0\%  & 12.53 & \cellcolor{GoogleRed}\textcolor{white}{\textbf{\$9.39M}} \\

\midrule

\multicolumn{9}{l}{\textbf{Grok 4.1 Fast}} \\
\midrule
\quad Baseline                      & 26.0\% & 63.76 & \$39.42M & \cellcolor{GoogleRed}\textcolor{white}{\textbf{100.0\%}} & 42.0\% & 28.0\% & 0.47  & \$27.24M \\
\quad RAP~\citep{kagaya2024rap}              & 28.0\% & 64.08 & \$35.47M & \cellcolor{GoogleRed}\textcolor{white}{\textbf{100.0\%}} & 40.0\% & 28.0\% & 0.41  & \$21.72M \\
\quad MemoryBank~\citep{zhong2024memorybank}         & 16.0\% & 58.24 & \$31.95M & \cellcolor{GoogleRed}\textcolor{white}{\textbf{100.0\%}} & 42.0\% & 20.0\% & 1.03  & \$11.65M \\
\quad Procedural Graph (Ours)           & \cellcolor{GoogleRed}\textcolor{white}{\textbf{40.0\%}} & \cellcolor{GoogleRed}\textcolor{white}{\textbf{75.14}} & \cellcolor{GoogleRed}\textcolor{white}{\textbf{\$39.62M}} & \cellcolor{GoogleRed}\textcolor{white}{\textbf{100.0\%}} & \cellcolor{GoogleRed}\textcolor{white}{\textbf{50.0\%}} & \cellcolor{GoogleRed}\textcolor{white}{\textbf{40.0\%}} & \cellcolor{GoogleRed}\textcolor{white}{\textbf{0.40}}  & \cellcolor{GoogleRed}\textcolor{white}{\textbf{\$30.11M}} \\

\bottomrule
\end{tabular}
\end{table*}

Figure~\ref{fig:cfo_ensemble_trajectories} compares cash trajectories and Kaplan--Meier survival curves for four planning configurations on Claude Sonnet 4.6, Gemini 3.1 Pro, Gemini 3.5 Flash, and Grok 4.1 Fast. Each model panel places survival curves above cash trajectories, which show individual sample paths, the mean, and the $95\%$ confidence interval. Survival gains vary across the four solvers.

The unguided baseline reaches full-horizon survival rates of $44.0\%$ for Claude Sonnet 4.6, $6.0\%$ for Gemini 3.1 Pro, $26.0\%$ for Grok 4.1 Fast, and $0.0\%$ for Gemini 3.5 Flash (Table~\ref{tab:gemini_grok_simulation}). In the Flash baseline, mean lifespan is $33.58$ months and no episode reaches the second crisis.

The Procedural Graph configuration (solid blue line) improves full-horizon survival for three solvers and mean lifespan for all four:

\begin{itemize}[leftmargin=*]
    \item \textbf{Claude Sonnet 4.6}: PG achieves the highest survival rate of 58.0\% (a 14.0-point gain over the baseline). Its mean cash trajectory remains stable and ends at \$76.03M, well above the liquidity warning threshold of \$5.0M.
    \item \textbf{Gemini 3.1 Pro}: The Procedural Graph provides the largest absolute survival gain, raising the survival rate from 6.0\% (baseline) to 34.0\%, and securing a final mean cash of \$22.99M, approximately $10.5$ times the baseline's \$2.18M.
    \item \textbf{Grok 4.1 Fast}: Survival improves from 26.0\% to 40.0\%, with the final mean cash increasing from \$23.62M to \$30.20M.
    \item \textbf{Gemini 3.5 Flash}: All four configurations have $0.0\%$ full-horizon survival in this evaluation, while PG increases mean lifespan from $33.58$ to $40.62$ months. This result describes the configurations evaluated here; the separate self-evolution experiment is reported in Section~\ref{subsec:evolution_results}.
\end{itemize}

MemoryBank achieves higher full-horizon survival than RAP on Gemini 3.1 Pro ($22.0\%$ vs.\ $6.0\%$) and Claude Sonnet 4.6 ($56.0\%$ vs.\ $50.0\%$). PG exceeds both on these two solvers and on Grok 4.1 Fast, while all methods tie at $0.0\%$ on Gemini 3.5 Flash.

Relative to the unguided baseline, PG raises full-horizon survival by $28$ points on Gemini 3.1 Pro and $14$ points on both Claude Sonnet 4.6 and Grok 4.1 Fast. Appendix~\ref{app:cfo_traces} examines the actions taken in one matched task instance.

\subsection{Case Study: Three Agents Facing the First Crisis}
\label{app:cfo_traces}

To understand the mechanisms driving the quantitative performance gap, we contrast the step-by-step reasoning traces of a baseline agent, a memory summarization agent, and a Procedural Graph guided agent. The following traces are extracted directly from real experimental logs of \texttt{Grok 4.1 Fast} operating on an identical task instance (Seed 14) during the critical transition into the first macroeconomic crisis (Months 29 to 33). Traces are condensed from the raw logs; bracketed italic text is editorial annotation, not model output.

In this baseline trace, the agent reports no bankruptcy risk despite declining cash reserves. It later submits requests that violate the environment's constraints:

\begin{tracebox}{GoogleRed}{Baseline Agent Trace (Sample 14) -- Myopia \& Rule Violation}
\footnotesize\raggedright
\textbf{Step 45 (Month 31):} \\*
\textbf{Thought:} Month 31 confirms trajectory: cash \$14.2M (down $\sim$\$7.4M but \textcolor{red}{\textbf{infinite runway $>$12mos}}) ... \textcolor{red}{\textbf{no bankruptcy risk}}. With 1 action left, \texttt{book\_closing} to advance month, lock YTD financials, sustain momentum, and maximize valuation... \\
\textbf{Action:} \texttt{book\_closing()} 

\smallskip
\textbf{Step 46 (Month 32):} \\*
\textbf{Thought:} Month 32: cash \$6.2M (down sharply $\sim$\$8M, runway $\sim$2-3mos, bankruptcy risk emerging)... Prior \texttt{book\_closing} advanced month but cash burn accelerating; cannot risk another \texttt{book\_closing} or pass. Raise equity to bolster cash... \\
\textbf{Action:} \texttt{fund\_raising\_request(type='equity', amount=20000000)} 

\smallskip
\textbf{Step 47 (Month 33):} \\*
\textbf{Thought:} Month 33: cash critically low at \$683K (runway $<$1mo, imminent bankruptcy risk)... Equity raise (\$20M requested, pending 1-6mos) submitted but not received; cannot pass or \texttt{book\_closing}. Submit debt raise for immediate bridge funding to avoid bankruptcy... \\
\textbf{Action:} \texttt{fund\_raising\_request(type='debt', amount=10000000)} \\
\textcolor{red}{\textbf{Observation: Your fundraising request was rejected: you already have a pending fundraising request.}}

\smallskip
\emph{[Observed behavior: The agent requests debt financing while an equity request is still pending. The environment rejects the additional request under the single-pending-request constraint.]}
\end{tracebox}

The memory summarization agent uses an LLM-generated running summary of past events. In this example, it recognizes the shrinking cash runway but submits fundraising amounts inconsistent with its stated intentions and repeats requests while earlier funding is still pending:

\begin{tracebox}{GoogleYellow}{Memory Summarization Agent (Sample 14) -- Hallucination \& Constraint Blindness}
\footnotesize\raggedright
\textbf{Step 51 (Month 29):} \\*
\textbf{Thought:} Cash \$25.3M (MoM -\$5.4M, runway $\sim$4mo per updated forecast)... pending equity (\$18M, submitted M26, now $\sim$3mo in)... Runway critically eroding (4mo $<$ 6mo buffer); pending raise may arrive soon but unreliable. Check market for raise viability... \\
\textbf{Action:} \texttt{check\_market\_data(data\_type='VIX')} 

\smallskip
\textbf{Step 53 (Month 30):} \\*
\textbf{Thought:} Cash \$25.3M (runway 4mo)... VIX 10.51 very low... Submit equity raise now per rules (prefer equity \$15-20M)... \\
\textbf{Action:} \texttt{fund\_raising\_request(type='equity', amount=}\textcolor{red}{\textbf{18}}\texttt{)} \\
\textcolor{red}{\textbf{Observation: Your fundraising request was rejected: you already have a pending fundraising request.}}

\smallskip
\textbf{Step 57 (Month 32):} \\*
\textbf{Thought:} Cash \$11.8M (runway 2mo)... low-risk equity raise \$18M fits rules... submit to extend runway... \\
\textbf{Action:} \texttt{fund\_raising\_request(type='equity', amount=}\textcolor{red}{\textbf{18}}\texttt{)} 

\smallskip
\textbf{Step 59 (Month 32):} \\*
\textbf{Thought:} Cash \$4.3M... runway 1mo critical... must raise now despite pending... submit \$20M equity to buffer urgency... \\
\textbf{Action:} \texttt{fund\_raising\_request(type='equity', amount=}\textcolor{red}{\textbf{20}}\texttt{)} \\
\textcolor{red}{\textbf{Observation: Your fundraising request was rejected: you already have a pending fundraising request.}}

\smallskip
\emph{[Observed behavior: The agent submits amounts of \$18 and \$20 despite referring to \$18M and \$20M in its reasoning, and repeatedly submits requests while earlier funding is pending.]}
\end{tracebox}

Guided by the Procedural Graph, the proposed agent maintains a clear structural prior of the task's temporal rules. It initiates fundraising with the correct scale well in advance, chaining sequential reasoning steps while remaining strictly compliant with environment constraints during the stochastic delivery lag:

\begin{tracebox}{GoogleGreen}{Procedural Graph Agent Trace (Sample 14) -- Strategic Foresight \& Adherence}
\footnotesize\raggedright
\textit{[At the opening of Month 30, an equity request submitted in Month 27 delivers: \$28.8M after the market-capacity cap.]}

\smallskip
\textbf{Step 37 (Month 30):} \\*
\textbf{Thought:} Equity raise succeeded (capped at $\sim$\$28.8M, added to cash for \$47.6M total)... VIX at 10.51 is exceptionally low (indicating calm markets, ideal for equity raise to minimize dilution cost)... \textcolor{GoogleGreen}{\textbf{with escalating burn ahead of the projected contraction, raise equity again now for $\sim$\$50M}}, preserving no-debt status... \\
\textbf{Action:} \texttt{fund\_raising\_request(type='equity', amount=50000000)}

\smallskip
\textbf{Step 38 (Month 30):} \\*
\textbf{Thought:} Request submitted (pending, 1--6 month delivery)... run forecast to confirm new runway... \\
\textbf{Action:} \texttt{cash\_flow\_forecast\_calculation(months=12)}

\smallskip
\textbf{Step 42 (Month 32):} \\*
\textbf{Thought:} ... forecast confirms 6-mo runway... pending \$50M equity (submitted M30) likely delivers soon... \textcolor{GoogleGreen}{\textbf{advance month via \texttt{book\_closing} to progress raise delivery}}... \\
\textbf{Action:} \texttt{book\_closing()}

\smallskip
\textbf{Step 44 (Month 33):} \\*
\textbf{Thought:} ... pending \$50M equity (submitted M30, now 3 mos in) likely delivers soon... no bankruptcy imminent... \textcolor{GoogleGreen}{\textbf{advance via \texttt{book\_closing} to progress raise delivery}}... \\
\textbf{Action:} \texttt{book\_closing()}

\smallskip
\emph{[Result: The agent \textcolor{GoogleGreen}{\textbf{strictly adheres to the delivery-lag and single-pending constraints, successfully bridges the crisis months, and survives}} to the end of the simulation at Month 132.]}
\end{tracebox}

In this task instance, the PG-guided agent tracks pending funding, checks projected runway, and advances the month while awaiting delivery. The baseline and memory-summary traces show invalid requests. These observations illustrate different behaviors on the same instance; they do not establish how frequently each behavior occurs across runs.

\section{Procedural Graph Construction: Supplementary Results}
\label{app:procedural_graph_construction}
This appendix provides quantitative results, diagnostic case studies of construction failure modes, formulations of the five PG construction strategies, and the detailed experimental configurations.

\subsection{Efficiency and Robustness Statistics}
\label{app:efficiency}
Tables~\ref{tab:hq_efficiency} (HotpotQA) and~\ref{tab:mc_efficiency} (MultiChallenge) compare task performance, resource use, and parsing failures across the five construction modes and the unguided baseline.
\begin{table}[htbp]
\centering
\scriptsize
\setlength{\aboverulesep}{0pt}
\setlength{\belowrulesep}{0pt}
\renewcommand{\arraystretch}{1.2}
\setlength{\tabcolsep}{10.5pt}
\caption{\textbf{HotpotQA test set performance, resource, and robustness statistics.} Highlighted cells mark column optima across all configurations, including the unguided baseline; ties are highlighted equally.}
\label{tab:hq_efficiency}
\begin{tabular}{l c c c c c}
\toprule
\rowcolor{GoogleBlue!15}
\textbf{Construction Mode} & \textbf{Ans F1 ($\uparrow$)} & \textbf{Tokens ($\downarrow$)} & \textbf{Steps ($\downarrow$)} & \textbf{Parsing Fail. ($\downarrow$)} & \textbf{Latency (s, $\downarrow$)} \\
\midrule
Unguided Baseline (w/o PG) & 71.21 & \cellcolor{GoogleBlue}\textcolor{white}{\textbf{4,003.24}} & 4.88 & 0.016 & \cellcolor{GoogleBlue}\textcolor{white}{\textbf{18.06}} \\
Mode 1: Hand-crafted Expert & 76.61 & 9,045.69 & 4.12 & 0.007 & 39.34 \\
Mode 2: Expert + Static Update & 77.16 & 8,942.79 & 4.11 & \cellcolor{GoogleBlue}\textcolor{white}{\textbf{0.005}} & 37.73 \\
Mode 3: Expert + Online Evolution & 76.34 & 10,658.36 & 4.11 & 0.009 & 39.61 \\
Mode 4: Scratch + Static Build & 69.49 & 6,396.90 & 4.30 & 0.016 & 36.98 \\
Mode 5: Scratch + Online Evolution & \cellcolor{GoogleBlue}\textcolor{white}{\textbf{78.79}} & 10,115.89 & \cellcolor{GoogleBlue}\textcolor{white}{\textbf{3.97}} & 0.016 & 31.53 \\
\bottomrule
\end{tabular}
\end{table}

\begin{table}[htbp]
\centering
\scriptsize
\setlength{\aboverulesep}{0pt}
\setlength{\belowrulesep}{0pt}
\renewcommand{\arraystretch}{1.2}
\setlength{\tabcolsep}{7.8pt}
\caption{\textbf{MultiChallenge test set performance, resource, and robustness statistics.} Bold blue cells mark column optima across all configurations, including the unguided baseline; ties are highlighted equally. Pale blue marks Mode 5's second-best values where Mode 2 ranks first.}
\label{tab:mc_efficiency}
\begin{tabular}{l c c c c c}
\toprule
\rowcolor{GoogleBlue!15}
\textbf{Construction Mode} & \textbf{Overall Success ($\uparrow$)} & \textbf{Tokens ($\downarrow$)} & \textbf{Steps ($\downarrow$)} & \textbf{Parsing Fail. ($\downarrow$)} & \textbf{Latency (s, $\downarrow$)} \\
\midrule
Unguided Baseline (w/o PG) & 87.50 & 7,403.98 & 7.05 & 1.05 & 127.59 \\
Mode 1: Hand-crafted Expert & 58.93 & 11,039.82 & 6.60 & 0.88 & 117.70 \\
Mode 2: Expert + Static Update & 53.57 & \cellcolor{GoogleBlue}\textcolor{white}{\textbf{5,990.79}} & \cellcolor{GoogleBlue}\textcolor{white}{\textbf{4.32}} & \cellcolor{GoogleBlue}\textcolor{white}{\textbf{0.36}} & \cellcolor{GoogleBlue}\textcolor{white}{\textbf{65.93}} \\
Mode 3: Expert + Online Evolution & \cellcolor{GoogleBlue}\textcolor{white}{\textbf{92.86}} & 12,157.52 & 6.73 & 0.98 & 128.50 \\
Mode 4: Scratch + Static Build & 89.29 & 14,859.80 & 8.02 & 1.07 & 166.57 \\
Mode 5: Scratch + Online Evolution & 91.07 & 7,984.50 & \cellcolor{GoogleBlue!15}5.05 & \cellcolor{GoogleBlue!15}0.57 & \cellcolor{GoogleBlue!15}92.81 \\
\bottomrule
\end{tabular}
\end{table}

\paragraph{Solver Steps and Latency.}
On MultiChallenge, Mode 5 uses $5.05$ steps and $92.81$ seconds per sample, compared with $8.02$ steps and $166.57$ seconds for Mode 4. Mode 2 has the lowest average step count ($4.32$) and latency ($65.93$ seconds), but also the lowest success rate ($53.57\%$). Mode 3 achieves the highest success rate ($92.86\%$), with $6.73$ steps and $128.50$ seconds per sample. These results show why resource use should be read alongside task performance; aggregate statistics alone do not identify the graph edits responsible for the differences.

Compared with the hand-crafted Mode 1 graph ($6.60$ steps and $117.70$ seconds), Mode 5 uses fewer steps and less time on MultiChallenge. On HotpotQA, Mode 5 has the lowest latency among the PG configurations ($31.53$ seconds).
\paragraph{Action Formatting and Parsing Robustness.}
On HotpotQA, expert-initialized Modes 1--3 have fewer parsing failures per sample than scratch-initialized Modes 4--5. Mode 2 has the lowest value ($0.005$), followed by Mode 1 ($0.007$) and Mode 3 ($0.009$); Modes 4 and 5 each have $0.016$. On MultiChallenge, Mode 5 records $0.57$ parsing failures per sample compared with Mode 4's $1.07$. The association between initialization and formatting therefore differs across tasks, and does not by itself establish that a particular graph schema causes the reduction.
\paragraph{Token Use.}
On MultiChallenge, Modes 1 and 3 use approximately $11.0\text{k}$ and $12.2\text{k}$ tokens per sample, respectively. Mode 2 uses approximately $6.0\text{k}$, alongside its lower success rate. Mode 5 uses $7{,}984.50$ tokens, compared with $14{,}859.80$ for Mode 4 and $7{,}403.98$ for the unguided baseline. The Mode 4--5 comparison reverses on HotpotQA: Mode 5 uses $10{,}115.89$ tokens versus Mode 4's $6{,}396.90$. Total token use reflects the complete interaction, including the number of steps and generated outputs, and cannot alone establish that a graph representation is more compact.

\subsection{Graph Construction Modes}
\label{app:mode_details}
We provide the formulation and implementation details for each of the five PG construction modes. Here, online evolution refers to incremental graph updates between training batches; the graph remains fixed within each episode and during test evaluation.
\begin{modebox}{Mode 1: Hand-crafted Expert PG (Zero-shot Baseline)}
\footnotesize\raggedright
\textbf{Initialization:} A human-engineered directed graph $G_{expert} = (V, E)$ containing expert-designed tool nodes, valid transitions, and manually annotated natural language guidance. \\
\textbf{Training Strategy:} None. Zero-shot execution on the test set. \\
\textbf{Evolutionary Mutation:} None. The graph structure and guidance remain static. \\
\textbf{Validation \& Safeguard:} None.
\end{modebox}
\begin{modebox}{Mode 2: Expert PG + Static One-time Update (Offline)}
\footnotesize\raggedright
\textbf{Refinement Mode:} \texttt{static\_onetime}. \\
\textbf{Initialization:} Initialized with the hand-crafted expert graph $G_{expert}$ from Mode 1. \\
\textbf{Training Strategy:} Static offline execution. The agent runs on the entire training split in a single pass to collect all successful and failed trajectories. \\
\textbf{Evolutionary Mutation:} Single-pass. The LLM refiner ingests all trajectories in a single large-context window to perform a one-time global offline update to transitions and guidance. \\
\textbf{Validation \& Safeguard:} None. The refined graph is directly committed.
\end{modebox}
\begin{modebox}{Mode 3: Expert PG + Online Evolution (Incremental)}
\footnotesize\raggedright
\textbf{Refinement Mode:} \texttt{static\_incremental}. \\
\textbf{Initialization:} Initialized with the hand-crafted expert graph $G_{expert}$ from Mode 1. \\
\textbf{Training Strategy:} Online incremental batches. The training split is partitioned into sequential strides ($S=100$ samples for HotpotQA; $S=20$ samples for MultiChallenge). \\
\textbf{Evolutionary Mutation:} After each stride, the refiner uses the latest failure logs to update nodes, transitions, and local guidance. \\
\textbf{Validation \& Safeguard:} The candidate is evaluated on the validation split. If performance declines, the previous best graph is restored.
\end{modebox}
\begin{modebox}{Mode 4: Scratch + Static One-time Build (Offline)}
\footnotesize\raggedright
\textbf{Refinement Mode:} \texttt{scratch\_onetime}. \\
\textbf{Initialization:} Minimal skeleton graph $G_{skeleton} = (\text{Start} \rightarrow \text{End})$ with no intermediate nodes or additional transitions. \\
\textbf{Training Strategy:} Static offline execution. The agent runs on the entire training split in a single pass to collect all successful and failed trajectories. \\
\textbf{Evolutionary Mutation:} Single-pass. The LLM refiner ingests all trajectories in a single large-context window to perform a one-time global offline update to transitions and guidance. \\
\textbf{Validation \& Safeguard:} None. The refined graph is directly committed.
\end{modebox}
\begin{modebox}{Mode 5: Scratch + Online Evolution (Incremental)}
\footnotesize\raggedright
\textbf{Refinement Mode:} \texttt{scratch\_incremental}. \\
\textbf{Initialization:} Minimal skeleton graph $G_{skeleton} = (\text{Start} \rightarrow \text{End})$ with no intermediate nodes or additional transitions. \\
\textbf{Training Strategy:} Online incremental batches. The training split is partitioned into sequential strides ($S=100$ samples for HotpotQA; $S=20$ samples for MultiChallenge). \\
\textbf{Evolutionary Mutation:} After each stride, the refiner uses the latest failure logs to update nodes, transitions, and local guidance. \\
\textbf{Validation \& Safeguard:} The candidate is evaluated on the validation split. If performance declines, the previous best graph is restored.
\end{modebox}

\subsection{Experimental Setup}
\label{app:construction_setup}
To ensure reproducibility, we detail the core experimental settings and dataset splits below:
\paragraph{Model Selection and Hyperparameters.}
All construction-strategy experiments in this appendix, including agent execution and evolutionary refinement, are powered by the frozen \texttt{Gemini 3.5 Flash} model snapshot. To ensure deterministic and reproducible tool-calling reasoning paths, the sampling temperature is set strictly to $0$ and top-$k$ sampling is configured with $k = 1$, enforcing greedy decoding across all model calls. Additionally, to prevent random execution deviations, the API safety filtering thresholds are unified and held constant to eliminate premature trajectory blocking, and the maximum generation length is locked at $2,048$ tokens for the agent and $8,192$ tokens for the evolutionary refiner to guard against truncated reasoning paths.
\paragraph{HotpotQA Environment Setup.}
The HotpotQA experiments are conducted on a standard multi-hop reasoning split consisting of $1,000$ training samples, $1,000$ validation samples, and $1,000$ test samples. The agent's performance is measured using strict string Exact Match (EM) and word-level overlap F1 score between the agent's generated answer and the ground truth.
\paragraph{MultiChallenge Environment Setup.}
The MultiChallenge dialogue experiments are conducted on a dataset split consisting of $100$ training samples, $100$ validation samples, and $56$ test samples. Task performance is evaluated via an LLM judge based on Gemini 3.1 Pro, measuring the Overall Success Rate, alongside four core challenge subcategory axes: Inference Memory (IM), Instruction Retention (IR), Reliable Versioned Editing (RVE), and Self Coherence (SC).

\section{Round-by-Round Self-Evolution on EnterpriseArena}
\label{app:self_evolution}

Table~\ref{tab:evolution_rounds} presents the round-by-round metrics of the $10$-generation CFO evolution. The delta ($\Delta$) rows track metric changes relative to the active best validation checkpoint. Appendix~\ref{app:evolution_efficiency} examines tool use, and Appendix~\ref{app:gate_rollbacks} discusses candidate screening.

\begin{table*}[!htb]
\centering
\scriptsize
\setlength{\aboverulesep}{0pt}
\setlength{\belowrulesep}{0pt}
\renewcommand{\arraystretch}{1} 
\setlength{\tabcolsep}{3.5pt}
\caption{\textbf{PG self-evolution across ten rounds with Gemini 3.5 Flash.} The $\Delta$ rows report metric changes relative to the previous best validation checkpoint. Gray rows mark rounds without a committed graph update.}
\label{tab:evolution_rounds}
\begin{tabular}{l c c c c c c c c c}
\toprule
\rowcolor{GoogleRed!15}
\cellcolor{GoogleRed!15} & \multicolumn{3}{c}{\textbf{Overall}} & \multicolumn{3}{c}{\textbf{Multi-Crisis Survival}} & \multicolumn{3}{c}{\textbf{Agent Performance}} \\

\arrayrulecolor{white}
\cmidrule(lr){2-4} \cmidrule(lr){5-7} \cmidrule(lr){8-10}
\arrayrulecolor{black}

\rowcolor{GoogleRed!15}
\multirow{-2}{*}{\textbf{Evolution Round}} & \textbf{Full Surv.\%} & \textbf{Avg. Months} & \textbf{Avg. Score} & \textbf{1st Crisis} & \textbf{2nd Crisis} & \textbf{3rd Crisis} & \textbf{Tools/Mo} & \textbf{Actions} & \textbf{Raised} \\
\midrule

\multicolumn{10}{l}{\textbf{Baseline}} \\
\quad Validate                  & 0.0\% & 34.80 & \$28.377M & 100.0\% & 0.0\% & 0.0\% & 17.23 & 35.7 & \$0.47M \\
\quad Test                      & 0.0\% & 33.30 & \$28.754M & 100.0\% & 0.0\% & 0.0\% & 18.60 & 34.2 & \$0.00M \\
\midrule

\multicolumn{10}{l}{\textbf{Round 01} \textcolor{GoogleGreen!65!black}{\textbf{[$\checkmark$ Evolved]}}} \\
\quad Train                     & 0.0\% & 33.40 & \$28.777M & 100.0\% & 0.0\% & 0.0\% & 16.25 & 34.4 & \$0.00M \\
\quad Validate                  & 45.0\% & 88.90 & \$38.594M & 100.0\% & 60.0\% & 45.0\% & 7.92 & 89.9 & \$40.68M \\
\quad \quad \textcolor{black!65}{\textit{\scriptsize $\Delta$ vs. Base}} & \textcolor{GoogleGreen!65!black}{\scriptsize +45.0} & \textcolor{GoogleGreen!65!black}{\scriptsize +54.10} & \textcolor{GoogleGreen!65!black}{\scriptsize +\$10.217M} & \textcolor{black!65}{\scriptsize --} & \textcolor{GoogleGreen!65!black}{\scriptsize +60.0} & \textcolor{GoogleGreen!65!black}{\scriptsize +45.0} & \textcolor{GoogleGreen!65!black}{\scriptsize -9.31} & \textcolor{GoogleGreen!65!black}{\scriptsize +54.2} & \textcolor{GoogleGreen!65!black}{\scriptsize +\$40.21M} \\
\quad Test                      & \textbf{70.0\%} & \textbf{104.15} & \textbf{\$43.519M} & 100.0\% & \textbf{80.0\%} & \textbf{70.0\%} & \textbf{3.62} & 105.2 & \textbf{\$44.90M} \\
\midrule

\multicolumn{10}{l}{\textbf{Round 02} \textcolor{GoogleGreen!65!black}{\textbf{[$\checkmark$ Evolved]}}} \\
\quad Train                     & 85.0\% & 117.75 & \$46.524M & 100.0\% & 90.0\% & 85.0\% & 3.39 & 118.8 & \$57.27M \\
\quad Validate                  & 80.0\% & 112.60 & \$46.038M & 100.0\% & 85.0\% & 80.0\% & 3.08 & 113.6 & \$52.33M \\
\quad \quad \textcolor{black!65}{\textit{\scriptsize $\Delta$ vs. R01}} & \textcolor{GoogleGreen!65!black}{\scriptsize +35.0} & \textcolor{GoogleGreen!65!black}{\scriptsize +23.70} & \textcolor{GoogleGreen!65!black}{\scriptsize +\$7.444M} & \textcolor{black!65}{\scriptsize --} & \textcolor{GoogleGreen!65!black}{\scriptsize +25.0} & \textcolor{GoogleGreen!65!black}{\scriptsize +35.0} & \textcolor{GoogleGreen!65!black}{\scriptsize -4.84} & \textcolor{GoogleGreen!65!black}{\scriptsize +23.7} & \textcolor{GoogleGreen!65!black}{\scriptsize +\$11.65M} \\
\quad Test                      & \textbf{80.0\%} & \textbf{114.00} & \textbf{\$44.992M} & 100.0\% & \textbf{90.0\%} & \textbf{80.0\%} & \textbf{3.11} & 115.0 & \textbf{\$46.83M} \\
\midrule

\rowcolor{black!4}
\multicolumn{10}{l}{\textbf{Round 03} \textcolor{GoogleRed!75!black}{\textbf{[$\times$ Rolled Back]}}} \\
\rowcolor{black!4}
\quad \textcolor{black!65}{Train}   & \textcolor{black!65}{80.0\%} & \textcolor{black!65}{113.35} & \textcolor{black!65}{\$46.857M} & \textcolor{black!65}{100.0\%} & \textcolor{black!65}{85.0\%} & \textcolor{black!65}{80.0\%} & \textcolor{black!65}{3.08} & \textcolor{black!65}{114.3} & \textcolor{black!65}{\$51.88M} \\
\rowcolor{black!4}
\quad \textcolor{black!65}{Validate}& \textcolor{black!65}{65.0\%} & \textcolor{black!65}{102.15} & \textcolor{black!65}{\$43.965M} & \textcolor{black!65}{100.0\%} & \textcolor{black!65}{75.0\%} & \textcolor{black!65}{70.0\%} & \textcolor{black!65}{3.30} & \textcolor{black!65}{103.2} & \textcolor{black!65}{\$46.74M} \\
\rowcolor{black!4}
\quad \quad \textcolor{black!65}{\textit{\scriptsize $\Delta$ vs. R02}} & \textcolor{GoogleRed!75!black}{\scriptsize -15.0} & \textcolor{GoogleRed!75!black}{\scriptsize -10.45} & \textcolor{GoogleRed!75!black}{\scriptsize -\$2.073M} & \textcolor{black!65}{\scriptsize --} & \textcolor{GoogleRed!75!black}{\scriptsize -10.0} & \textcolor{GoogleRed!75!black}{\scriptsize -10.0} & \textcolor{GoogleRed!75!black}{\scriptsize +0.22} & \textcolor{GoogleRed!75!black}{\scriptsize -10.4} & \textcolor{GoogleRed!75!black}{\scriptsize -\$5.59M} \\
\midrule

\rowcolor{black!4}
\multicolumn{10}{l}{\textbf{Round 04} \textcolor{GoogleRed!75!black}{\textbf{[$\times$ Rolled Back]}}} \\
\rowcolor{black!4}
\quad \textcolor{black!65}{Train}   & \textcolor{black!65}{80.0\%} & \textcolor{black!65}{115.05} & \textcolor{black!65}{\$46.631M} & \textcolor{black!65}{100.0\%} & \textcolor{black!65}{90.0\%} & \textcolor{black!65}{80.0\%} & \textcolor{black!65}{3.06} & \textcolor{black!65}{116.0} & \textcolor{black!65}{\$53.04M} \\
\rowcolor{black!4}
\quad \textcolor{black!65}{Validate}& \textcolor{black!65}{75.0\%} & \textcolor{black!65}{109.20} & \textcolor{black!65}{\$49.831M} & \textcolor{black!65}{100.0\%} & \textcolor{black!65}{85.0\%} & \textcolor{black!65}{75.0\%} & \textcolor{black!65}{3.05} & \textcolor{black!65}{110.2} & \textcolor{black!65}{\$70.85M} \\
\rowcolor{black!4}
\quad \quad \textcolor{black!65}{\textit{\scriptsize $\Delta$ vs. R02}} & \textcolor{GoogleRed!75!black}{\scriptsize -5.0} & \textcolor{GoogleRed!75!black}{\scriptsize -3.40} & \textcolor{GoogleGreen!65!black}{\scriptsize +\$3.793M} & \textcolor{black!65}{\scriptsize --} & \textcolor{black!65}{\scriptsize --} & \textcolor{GoogleRed!75!black}{\scriptsize -5.0} & \textcolor{GoogleGreen!65!black}{\scriptsize -0.03} & \textcolor{GoogleRed!75!black}{\scriptsize -3.4} & \textcolor{GoogleGreen!65!black}{\scriptsize +\$18.52M} \\
\midrule

\rowcolor{black!4}
\multicolumn{10}{l}{\textbf{Round 05}} \\
\rowcolor{black!4}
\quad \textcolor{black!65}{Train}   & \textcolor{black!65}{80.0\%} & \textcolor{black!65}{114.05} & \textcolor{black!65}{\$44.565M} & \textcolor{black!65}{100.0\%} & \textcolor{black!65}{90.0\%} & \textcolor{black!65}{80.0\%} & \textcolor{black!65}{3.12} & \textcolor{black!65}{115.0} & \textcolor{black!65}{\$52.85M} \\
\rowcolor{black!4}
\quad \textcolor{black!65}{Validate (R04)}& \textcolor{black!65}{75.0\%} & \textcolor{black!65}{109.20} & \textcolor{black!65}{\$49.831M} & \textcolor{black!65}{100.0\%} & \textcolor{black!65}{85.0\%} & \textcolor{black!65}{75.0\%} & \textcolor{black!65}{3.05} & \textcolor{black!65}{110.2} & \textcolor{black!65}{\$70.85M} \\
\midrule

\rowcolor{black!4}
\multicolumn{10}{l}{\textbf{Round 06} \textcolor{GoogleRed!75!black}{\textbf{[$\times$ Rolled Back]}}} \\
\rowcolor{black!4}
\quad \textcolor{black!65}{Train}   & \textcolor{black!65}{80.0\%} & \textcolor{black!65}{114.90} & \textcolor{black!65}{\$46.345M} & \textcolor{black!65}{100.0\%} & \textcolor{black!65}{90.0\%} & \textcolor{black!65}{80.0\%} & \textcolor{black!65}{3.09} & \textcolor{black!65}{115.9} & \textcolor{black!65}{\$56.52M} \\
\rowcolor{black!4}
\quad \textcolor{black!65}{Validate}& \textcolor{black!65}{65.0\%} & \textcolor{black!65}{108.75} & \textcolor{black!65}{\$43.831M} & \textcolor{black!65}{100.0\%} & \textcolor{black!65}{80.0\%} & \textcolor{black!65}{80.0\%} & \textcolor{black!65}{2.91} & \textcolor{black!65}{109.8} & \textcolor{black!65}{\$54.54M} \\
\rowcolor{black!4}
\quad \quad \textcolor{black!65}{\textit{\scriptsize $\Delta$ vs. R02}} & \textcolor{GoogleRed!75!black}{\scriptsize -15.0} & \textcolor{GoogleRed!75!black}{\scriptsize -3.85} & \textcolor{GoogleRed!75!black}{\scriptsize -\$2.207M} & \textcolor{black!65}{\scriptsize --} & \textcolor{GoogleRed!75!black}{\scriptsize -5.0} & \textcolor{black!65}{\scriptsize --} & \textcolor{GoogleGreen!65!black}{\scriptsize -0.17} & \textcolor{GoogleRed!75!black}{\scriptsize -3.8} & \textcolor{GoogleGreen!65!black}{\scriptsize +\$2.21M} \\
\midrule

\multicolumn{10}{l}{\textbf{Round 07} \textcolor{GoogleGreen!65!black}{\textbf{[$\checkmark$ Evolved]}}} \\
\quad Train                     & 65.0\% & 99.70 & \$40.722M & 100.0\% & 75.0\% & 65.0\% & 3.19 & 100.7 & \$40.46M \\
\quad Validate                  & 80.0\% & 114.15 & \$47.607M & 100.0\% & 90.0\% & 80.0\% & 3.12 & 115.2 & \$68.91M \\
\quad \quad \textcolor{black!65}{\textit{\scriptsize $\Delta$ vs. R02}} & \textcolor{black!65}{\scriptsize --} & \textcolor{GoogleGreen!65!black}{\scriptsize +1.55} & \textcolor{GoogleGreen!65!black}{\scriptsize +\$1.569M} & \textcolor{black!65}{\scriptsize --} & \textcolor{GoogleGreen!65!black}{\scriptsize +5.0} & \textcolor{black!65}{\scriptsize --} & \textcolor{GoogleRed!75!black}{\scriptsize +0.04} & \textcolor{GoogleGreen!65!black}{\scriptsize +1.6} & \textcolor{GoogleGreen!65!black}{\scriptsize +\$16.58M} \\
\quad Test                      & \textbf{95.0\%} & \textbf{126.05} & \textbf{\$47.360M} & 100.0\% & \textbf{95.0\%} & \textbf{95.0\%} & \textbf{3.10} & \textbf{127.0} & \textbf{\$68.94M} \\
\midrule

\multicolumn{10}{l}{\textbf{Round 08} \textcolor{GoogleGreen!65!black}{\textbf{[$\checkmark$ Evolved]}}} \\
\quad Train                     & 85.0\% & 120.35 & \$49.781M & 100.0\% & 90.0\% & 90.0\% & 3.12 & 121.3 & \$77.33M \\
\quad Validate                  & 90.0\% & 121.20 & \$57.371M & 100.0\% & 90.0\% & 90.0\% & 3.13 & 122.2 & \$94.32M \\
\quad \quad \textcolor{black!65}{\textit{\scriptsize $\Delta$ vs. R07}} & \textcolor{GoogleGreen!65!black}{\scriptsize +10.0} & \textcolor{GoogleGreen!65!black}{\scriptsize +7.05} & \textcolor{GoogleGreen!65!black}{\scriptsize +\$9.764M} & \textcolor{black!65}{\scriptsize --} & \textcolor{black!65}{\scriptsize --} & \textcolor{GoogleGreen!65!black}{\scriptsize +10.0} & \textcolor{GoogleRed!75!black}{\scriptsize +0.01} & \textcolor{GoogleGreen!65!black}{\scriptsize +7.0} & \textcolor{GoogleGreen!65!black}{\scriptsize +\$25.41M} \\
\quad Test                      & 85.0\% & 116.30 & \textbf{\$54.367M} & 100.0\% & 85.0\% & 85.0\% & 3.12 & 117.3 & \textbf{\$82.32M} \\
\midrule

\multicolumn{10}{l}{\textbf{Round 09} \textcolor{GoogleGreen!65!black}{\textbf{[$\checkmark$ Evolved]}}} \\
\quad Train                     & 95.0\% & 126.05 & \$58.188M & 100.0\% & 95.0\% & 95.0\% & 3.08 & 127.0 & \$97.84M \\
\quad Validate                  & 90.0\% & 121.20 & \$56.570M & 100.0\% & 90.0\% & 90.0\% & 3.08 & 122.2 & \$93.05M \\
\quad \quad \textcolor{black!65}{\textit{\scriptsize $\Delta$ vs. R08}} & \textcolor{black!65}{\scriptsize --} & \textcolor{black!65}{\scriptsize --} & \textcolor{GoogleRed!75!black}{\scriptsize -\$0.801M} & \textcolor{black!65}{\scriptsize --} & \textcolor{black!65}{\scriptsize --} & \textcolor{black!65}{\scriptsize --} & \textcolor{GoogleGreen!65!black}{\scriptsize -0.05} & \textcolor{black!65}{\scriptsize --} & \textcolor{GoogleRed!75!black}{\scriptsize -\$1.27M} \\
\quad Test                      & 85.0\% & 116.30 & \textbf{\$54.440M} & 100.0\% & 85.0\% & 85.0\% & \textbf{3.06} & 117.3 & \$82.32M \\
\midrule

\rowcolor{black!4}
\multicolumn{10}{l}{\textbf{Round 10} \textcolor{GoogleRed!75!black}{\textbf{[$\times$ Rolled Back]}}} \\
\rowcolor{black!4}
\quad \textcolor{black!65}{Train}   & \textcolor{black!65}{90.0\%} & \textcolor{black!65}{121.15} & \textcolor{black!65}{\$58.418M} & \textcolor{black!65}{100.0\%} & \textcolor{black!65}{90.0\%} & \textcolor{black!65}{90.0\%} & \textcolor{black!65}{3.07} & \textcolor{black!65}{122.2} & \textcolor{black!65}{\$88.32M} \\
\rowcolor{black!4}
\quad \textcolor{black!65}{Validate}& \textcolor{black!65}{85.0\%} & \textcolor{black!65}{121.20} & \textcolor{black!65}{\$56.386M} & \textcolor{black!65}{100.0\%} & \textcolor{black!65}{90.0\%} & \textcolor{black!65}{90.0\%} & \textcolor{black!65}{3.01} & \textcolor{black!65}{122.2} & \textcolor{black!65}{\$95.76M} \\
\rowcolor{black!4}
\quad \quad \textcolor{black!65}{\textit{\scriptsize $\Delta$ vs. R09}} & \textcolor{GoogleRed!75!black}{\scriptsize -5.0} & \textcolor{black!65}{\scriptsize --} & \textcolor{GoogleRed!75!black}{\scriptsize -\$0.184M} & \textcolor{black!65}{\scriptsize --} & \textcolor{black!65}{\scriptsize --} & \textcolor{black!65}{\scriptsize --} & \textcolor{GoogleGreen!65!black}{\scriptsize -0.07} & \textcolor{black!65}{\scriptsize --} & \textcolor{GoogleGreen!65!black}{\scriptsize +\$2.71M} \\
\bottomrule
\end{tabular}
\end{table*}

\subsection{Computational Efficiency}
\label{app:evolution_efficiency}
Beyond task performance, the evolved graphs also help reduce redundant tool use. As detailed in Table~\ref{tab:evolution_rounds}, the average number of tool invocations per simulation month (\texttt{Tools/Mo}) on the validation split drops from \textbf{$17.23$} in the Baseline to \textbf{$3.08$} in Round 2, stabilizing at \textbf{$3.13$} by Round 8.
The baseline agent, operating under zero-shot ReAct, repeatedly queries the environment for cash balances and market metrics within the same turn. The evolved PG guides the agent through these queries once per cycle. This is an \textbf{$81.8\%$ reduction in tool invocations} per simulation month, showing that the evolved graph reduces environment interactions in this experiment. We report tool calls rather than tokens here because the simulation logs record invocation counts but not per-month token consumption.

\subsection{Candidate Screening: Three Cases}
\label{app:gate_rollbacks}
The evolution loop combines validation checks with structural verification to filter unsuccessful candidates, as illustrated by three cases:
\begin{itemize}[leftmargin=*]
    \item \textbf{Performance Rollback (Rounds 3, 4, 6)}: In Round 3, the mutation engine attempted to lower the cash threshold that triggers fundraising. While this succeeded on specific training seeds, it led to premature dilution and cash shortages on validation seeds, dropping validation survival by \textbf{$15.0$ points} and average lifespan by \textbf{$10.45$ months}. The gate rejected the candidate and restored the Round 2 checkpoint.
    \item \textbf{Execution Constraint Rollback (Round 5)}: In Round 5, the candidate failed structural verification before simulation, so validation was skipped. The reported validation metrics for this round carry forward the Round 4 results.
    \item \textbf{Validation Safeguard (Round 10)}: In Round 10, training survival was $90.0\%$, but validation survival fell to $85.0\%$ (\textbf{$-5.0$} points) and the mean time-averaged enterprise score on the validation split decreased by \textbf{$\$0.184\text{M}$}. The validation gate rejected the mutation because its strong training performance did not carry over to validation, and the loop terminated with the Round 9 graph.
\end{itemize}
\subsection{Step-by-Step Topological Analysis}
\label{sec:app_topology}
\begin{figure}[!htbp]
\centering
\includegraphics[width=0.91\linewidth,trim=190bp 190bp 190bp 190bp,clip]{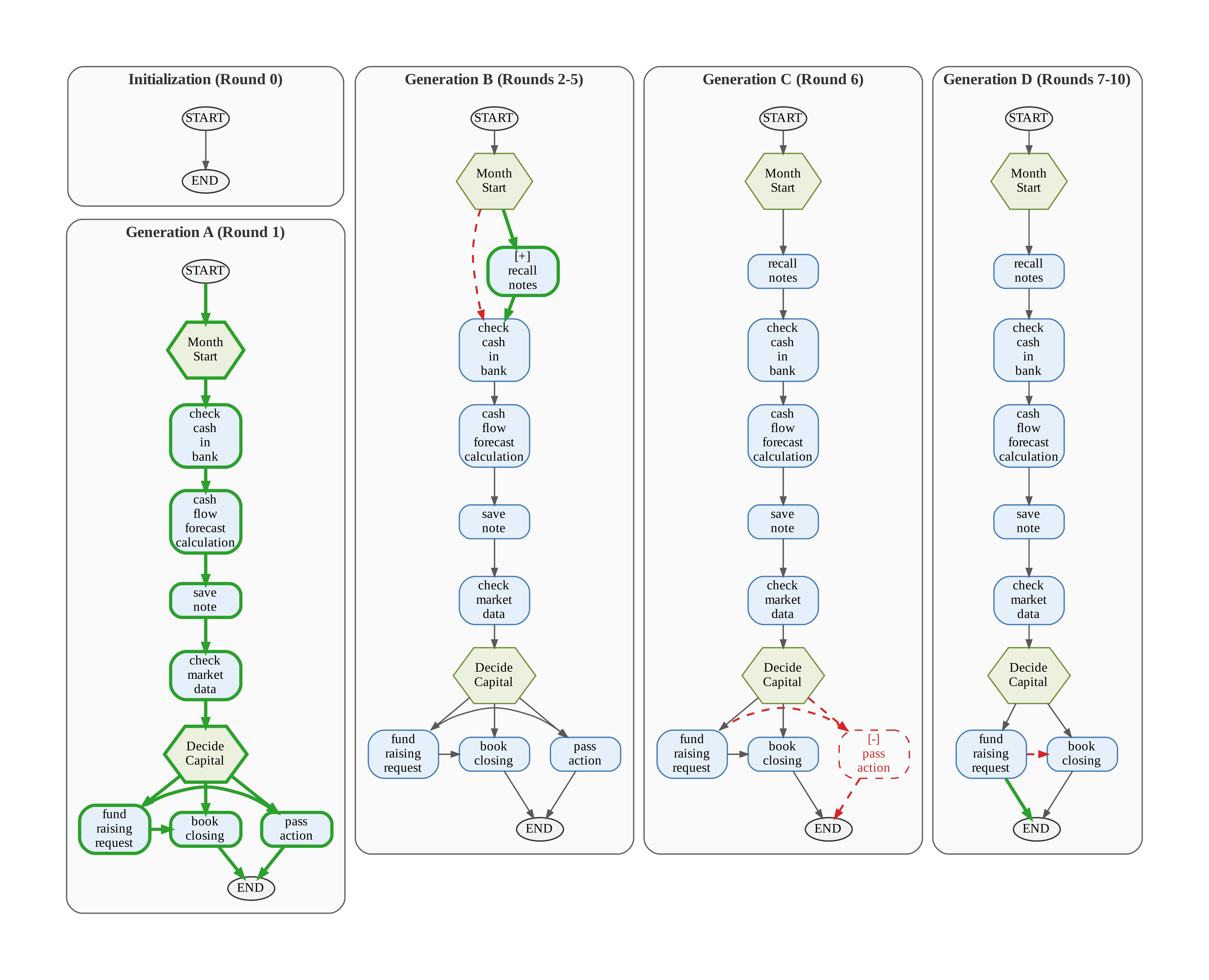}
\caption{\textbf{Topological evolution of the CFO Procedural Graph.} Green nodes/edges [+] denote additions; red dashed nodes/edges [-] denote deletions/pruning.}
\label{fig:pg_evolution}
\end{figure}

Figure~\ref{fig:pg_evolution} visualizes the topological changes of the CFO Procedural Graph across key evolutionary stages.

\begin{itemize}[leftmargin=*]
    \item \textbf{Initialization (Round 0 - Baseline)}: The agent has no structural prior ($\text{Start} \rightarrow \text{End}$). Without intermediate procedural guidance, the LLM must infer the action sequence from the running trajectory, leading to high computational cost, redundant tool calls, and missed tasks such as cash forecasts.
    \newpage
    \item \begin{samepage}\textbf{Generation A (Round 1 - Backbone Discovery)}: The loop suggests the following sequence: \texttt{Start} $\rightarrow$ \texttt{Month\_Start} $\rightarrow$ \texttt{check\_cash\_in\_bank} $\rightarrow$ \texttt{cash\_flow\_forecast\_calculation} $\rightarrow$ \texttt{save\_note} $\rightarrow$ \texttt{check\_market\_data} $\rightarrow$ \texttt{Decide\_Capital}. This structure guides the agent to audit cash, project runway, save context, and check market valuation before making a financing decision.\par\end{samepage}
    \item \textbf{Generation B (Rounds 2--6 - Working Memory)}: The loop inserts \texttt{recall\_notes} immediately after \texttt{Month\_Start}. By guiding the agent to use \texttt{save\_note} after the forecast and \texttt{recall\_notes} at the start of the next month, the Procedural Graph provides \textbf{durable external working memory} that preserves key metrics across months.
    \item \textbf{Generation C (Round 7 - Branch Pruning)}: The loop prunes the \texttt{pass\_action} node (the ``do nothing'' action) and its incident edges. The resulting graph retains the branches from \texttt{Decide\_Capital} to \texttt{fund\_raising\_request} and \texttt{book\_closing}. This topology is adopted in Round 7.
    \item \textbf{Generation D (Rounds 8--10 - Administrative Bypass)}: The edge from \texttt{fund\_raising\_request} to \texttt{book\_closing} is replaced with a direct link to \texttt{End} in Round 8, and this topology is retained through Round 10. Because the environment adapter already advances the month after a fundraising request, the revised graph suggests ending the current procedure and reassessing the new month's state instead of immediately following the request with another month-advancing action.
\end{itemize}

\subsection{Self-Evolution on HotpotQA and MultiChallenge}
\label{sec:app_cross_validation}
\begin{figure}[htbp]
\centering
\includegraphics[width=\textwidth]{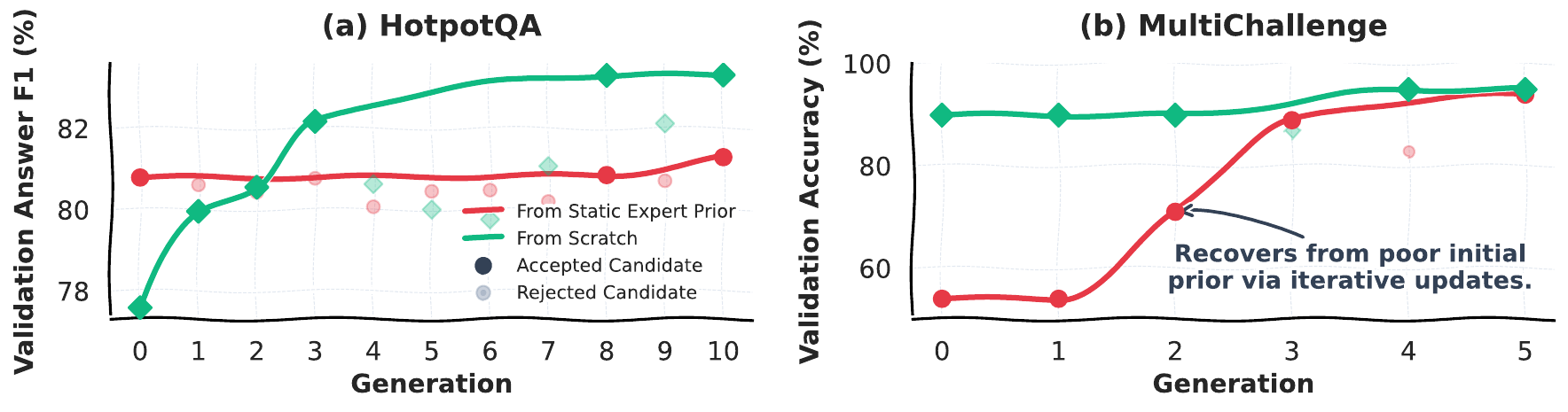}
\caption{\textbf{Self-evolution trajectory on (a) HotpotQA and (b) MultiChallenge.}}
\label{fig:cross_evolution}
\end{figure}

We also examine self-evolution on \textbf{HotpotQA} (multi-hop reasoning; Appendix~\ref{app:hotpotqa_evolution}) and \textbf{MultiChallenge} (complex instruction following; Appendix~\ref{app:multichallenge_evolution}), comparing \textbf{Mode 3} (evolving from an expert-designed prior) with \textbf{Mode 5} (evolving from scratch). Figure~\ref{fig:cross_evolution} shows the resulting trajectories.

\subsubsection{HotpotQA: Emergent Simplicity}
\label{app:hotpotqa_evolution}
In Figure~\ref{fig:cross_evolution}(a), Mode 5 starts from scratch with an F1 score of $77.59\%$, below Mode 3. It subsequently surpasses Mode 3 and reaches a peak validation F1 of $83.31\%$ at Generation 10.

These results show that, in our HotpotQA study, evolution from a minimal skeleton can yield more effective guidance than evolution from the hand-designed prior used here. The test-set statistics in Table~\ref{tab:hq_efficiency} also show that Mode 5 uses fewer solver steps than the expert-initialized modes. Its total token use remains higher than in Modes 1, 2, and 4, so shorter trajectories do not imply lower token cost.
\subsubsection{MultiChallenge: Prior Correction and Recovery}
\label{app:multichallenge_evolution}
Figure~\ref{fig:cross_evolution}(b) demonstrates the loop's robustness when initialized with a mismatched or poorly designed prior. The hand-designed graph used to initialize Mode 3 was unsuited for the MultiChallenge task, guiding the agent toward incorrect tool loops and yielding an initial validation accuracy of only \textbf{$54.0\%$}.
The self-evolution loop successfully repaired this prior through two phases:
\begin{enumerate}[leftmargin=*]
    \item \textbf{Pruning}: The mutation engine identified and deleted the mismatched transition edges, reducing the graph's edge count from $13$ to $10$ by Generation 5.
    \item \textbf{Restructuring}: The loop reconstructed the control flow to align with the task's constraints, raising validation accuracy to \textbf{$93.9\%$} (ultimately approaching the scratch-built Mode 5 at $94.9\%$).
\end{enumerate}
These results show that iterative evolution can recover from an unsuitable human-designed initialization.

\section{Additional Execution Cases}
\label{app:additional_examples}
The following cases use Gemini 3.5 Flash and examine stopping decisions on BFCL (Appendix~\ref{app:bfcl_stopping}) and graph revision on MultiChallenge (Appendix~\ref{app:multichallenge_edits}). Quotations come from solver traces or saved graph attributes, as labeled.

\subsection{BFCL: Stopping After a Requested Quote}
\label{app:bfcl_stopping}
In test sample 051, both agents receive the same request for an economy-class airfare and retrieve a \$220 quote. The PG agent uses a graph built from scratch by a one-time offline update and ends the turn after reporting the price; the baseline continues into payment and booking operations.

\begin{tracebox}{GoogleBlue}{BFCL: Requested Result and Stopping Behavior}
\footnotesize\raggedright
\textbf{Observed tool result, both runs:}\\
\texttt{get\_flight\_cost} returns \texttt{travel\_cost\_list: [220.0]}.

\smallskip
\textbf{PG solver rationale (verbatim excerpt):}\\
``Since the user did not explicitly request to book the flight, I should not proceed to book\_flight.''

\smallskip
\textbf{PG action (verbatim):}\\
\texttt{Finish(final\_response="The flight cost from Rivermist (RMS) to Stonebrook (SBK) on October 6, 2024, in economy class is \$220.00.")}

\smallskip
\textbf{Baseline continuation (editorial summary):}\\
The agent proceeds to authentication, card operations, and \texttt{book\_flight}. After an unsuccessful booking attempt, it changes the budget limit and books the flight. The environment reports a state mismatch at turn 0.
\end{tracebox}

The PG run advances to the next user instruction and eventually succeeds; the baseline fails on the first turn after taking actions beyond the requested quote.

\subsection{MultiChallenge: Correcting the Response Objective}
\label{app:multichallenge_edits}
Validation sample 059 asks for a joke about renewable energy or Sheffield under an earlier instruction to use only passive voice. The environment also exposes the target question, ``Did the model consistently use passive sentence construction?'' We compare Generation 2 and Generation 3 candidate evaluations from Mode 3; their initial query and full dialogue history are identical.

\begin{tracebox}{GoogleGreen}{MultiChallenge: Graph Revision and Observed Responses}
\footnotesize\raggedright
\textbf{Generation 2 candidate attribute (verbatim):}\\
\texttt{AnalyzeTargetQuestion} $\rightarrow$ \texttt{Finish}: ``Direct finish for simple evaluations without drafting.''

\smallskip
\textbf{Generation 2 recorded actions:}\\
\texttt{ParseHistory} $\rightarrow$ \texttt{AnalyzeTargetQuestion} $\rightarrow$ \texttt{Finish}.\\
\textbf{Final response excerpt:} ``No, the model did not consistently use passive sentence construction.''

\smallskip
\textbf{Generation 3 graph changes:}\\
The direct edge to \texttt{Finish} is removed. Guidance on \texttt{ExtractConstraints} $\rightarrow$ \texttt{Finish} now states: ``Do NOT write a meta-evaluation or answer the target question directly.''

\smallskip
\textbf{Generation 3 recorded actions:}\\
\texttt{ParseHistory} $\rightarrow$ \texttt{AnalyzeTargetQuestion} $\rightarrow$ \texttt{ExtractConstraints} $\rightarrow$ \texttt{Finish}.\\
\textbf{Final response (verbatim):} ``A joke about renewable energy is being shared. Why are wind turbines loved by everyone? Many fans are known to be made by them.''
\end{tracebox}

The saved success indicator changes from 0 to 1. The revised outgoing path and edge advice correspond to a shift from evaluating the dialogue to answering the user.


\clearpage
\begin{thebibliography}{54}
\providecommand{\natexlab}[1]{#1}
\providecommand{\url}[1]{\texttt{#1}}
\expandafter\ifx\csname urlstyle\endcsname\relax
  \providecommand{\doi}[1]{doi: #1}\else
  \providecommand{\doi}{doi: \begingroup \urlstyle{rm}\Url}\fi

\bibitem[Besta et~al.(2024)Besta, Blach, Kubicek, Gerstenberger, Podstawski,
  Gianinazzi, Gajda, Lehmann, Niewiadomski, Nyczyk, et~al.]{besta2024graph}
M.~Besta, N.~Blach, A.~Kubicek, R.~Gerstenberger, M.~Podstawski, L.~Gianinazzi,
  J.~Gajda, T.~Lehmann, H.~Niewiadomski, P.~Nyczyk, et~al.
\newblock Graph of thoughts: Solving elaborate problems with large language
  models.
\newblock In \emph{Proceedings of the AAAI conference on artificial
  intelligence}, volume~38, pages 17682--17690, 2024.

\bibitem[Deshpande et~al.(2025)Deshpande, Sirdeshmukh, Mols, Jin,
  Hernandez-Cardona, Lee, Kritz, Primack, Yue, and
  Xing]{deshpande2025multichallenge}
K.~Deshpande, V.~Sirdeshmukh, J.~B. Mols, L.~Jin, E.-Y. Hernandez-Cardona,
  D.~Lee, J.~Kritz, W.~E. Primack, S.~Yue, and C.~Xing.
\newblock {MultiChallenge}: A realistic multi-turn conversation evaluation
  benchmark challenging to frontier {LLMs}.
\newblock In \emph{Findings of the Association for Computational Linguistics:
  ACL 2025}, pages 18632--18702, 2025.

\bibitem[Du et~al.(2024)Du, Wei, and Zhang]{du2024anytool}
Y.~Du, F.~Wei, and H.~Zhang.
\newblock {AnyTool}: Self-reflective, hierarchical agents for large-scale {API}
  calls.
\newblock \emph{arXiv preprint arXiv:2402.04253}, 2024.

\bibitem[Edge et~al.(2024)Edge, Trinh, Cheng, Bradley, Chao, Mody, Truitt,
  Metropolitansky, Ness, and Larson]{edge2024local}
D.~Edge, H.~Trinh, N.~Cheng, J.~Bradley, A.~Chao, A.~Mody, S.~Truitt,
  D.~Metropolitansky, R.~O. Ness, and J.~Larson.
\newblock From local to global: A graph {RAG} approach to query-focused
  summarization.
\newblock \emph{arXiv preprint arXiv:2404.16130}, 2024.

\bibitem[Fang et~al.(2025)Fang, Liang, Wang, Wu, Qiao, Xie, Huang, Chen, and
  Zhang]{fang2025memp}
R.~Fang, Y.~Liang, X.~Wang, J.~Wu, S.~Qiao, P.~Xie, F.~Huang, H.~Chen, and
  N.~Zhang.
\newblock {Memp}: Exploring agent procedural memory.
\newblock \emph{arXiv preprint arXiv:2508.06433}, 2025.

\bibitem[Fu et~al.(2024)Fu, Kim, Kim, Sohn, Logeswaran, Bae, and
  Lee]{fu2024autoguide}
Y.~Fu, D.-K. Kim, J.~Kim, S.~Sohn, L.~Logeswaran, K.~Bae, and H.~Lee.
\newblock {AutoGuide}: Automated generation and selection of context-aware
  guidelines for large language model agents.
\newblock \emph{Advances in Neural Information Processing Systems},
  37:\penalty0 119919--119948, 2024.

\bibitem[Gao et~al.(2025)Gao, Wang, Peng, Tang, Shang, Sun, and
  Su]{gao2025tool}
L.~Gao, Y.~Wang, M.~Peng, J.~Tang, Y.~Shang, M.~Sun, and J.~Su.
\newblock Tool graph retriever: Exploring dependency graph-based tool retrieval
  for large language models.
\newblock \emph{arXiv preprint arXiv:2508.05152}, 2025.

\bibitem[Han et~al.(2026)Han, Wang, Qian, Li, Cao, He, Peng, Shen, Xu, Chen,
  et~al.]{han2026can}
Y.~Han, Y.~Wang, L.~Qian, H.~Li, Y.~Cao, Y.~He, X.~Peng, N.~Shen, Y.~Xu,
  Y.~Chen, et~al.
\newblock Can {LLM} agents be {CFOs}? benchmarking long-horizon resource
  allocation in an uncertain enterprise environment.
\newblock \emph{arXiv preprint arXiv:2603.23638}, 2026.

\bibitem[Huang et~al.(2024)Huang, Shi, Li, Fan, Wu, Zhang, Liu, Zhou, Wan,
  Gong, et~al.]{huang2024metatool}
Y.~Huang, J.~Shi, Y.~Li, C.~Fan, S.~Wu, Q.~Zhang, Y.~Liu, P.~Zhou, Y.~Wan,
  N.~Gong, et~al.
\newblock {MetaTool} benchmark for large language models: Deciding whether to
  use tools and which to use.
\newblock In \emph{International Conference on Learning Representations},
  volume 2024, pages 42978--43007, 2024.

\bibitem[Jiang et~al.(2025)Jiang, Zhou, GU, Han, and Li]{jiang2025naviagent}
Y.~Jiang, H.~Zhou, L.~GU, A.~Han, and T.~Li.
\newblock {NaviAgent}: Bilevel planning on tool navigation graph for
  large-scale orchestration.
\newblock \emph{arXiv preprint arXiv:2506.19500}, 2025.

\bibitem[Kagaya et~al.(2024)Kagaya, Yuan, Lou, Karlekar, Pranata, Kinose,
  Oguri, Wick, and You]{kagaya2024rap}
T.~Kagaya, T.~J. Yuan, Y.~Lou, J.~Karlekar, S.~Pranata, A.~Kinose, K.~Oguri,
  F.~Wick, and Y.~You.
\newblock {RAP}: Retrieval-augmented planning with contextual memory for
  multimodal {LLM} agents.
\newblock In \emph{NeurIPS 2024 Workshop on Open-World Agents}, 2024.

\bibitem[Li et~al.(2023)Li, Zhao, Yu, Song, Li, Yu, Li, Huang, and
  Li]{li2023api}
M.~Li, Y.~Zhao, B.~Yu, F.~Song, H.~Li, H.~Yu, Z.~Li, F.~Huang, and Y.~Li.
\newblock {API-Bank}: A comprehensive benchmark for tool-augmented {LLMs}.
\newblock In \emph{Proceedings of the 2023 conference on empirical methods in
  natural language processing}, pages 3102--3116, 2023.

\bibitem[Liu et~al.(2024{\natexlab{a}})Liu, Peng, Yi, Xie, Xiang, Liu, and
  Xu]{liu2024toolnet}
X.~Liu, Z.~Peng, X.~Yi, X.~Xie, L.~Xiang, Y.~Liu, and D.~Xu.
\newblock {ToolNet}: Connecting large language models with massive tools via
  tool graph.
\newblock \emph{arXiv preprint arXiv:2403.00839}, 2024{\natexlab{a}}.

\bibitem[Liu et~al.(2024{\natexlab{b}})Liu, Lai, Gao, Cui, Li, Zhu, Lu, Chen,
  Qiao, Dai, et~al.]{liu2024controlllm}
Z.~Liu, Z.~Lai, Z.~Gao, E.~Cui, Z.~Li, X.~Zhu, L.~Lu, Q.~Chen, Y.~Qiao, J.~Dai,
  et~al.
\newblock {ControlLLM}: Augment language models with tools by searching on
  graphs.
\newblock In \emph{European Conference on Computer Vision}, pages 89--105.
  Springer, 2024{\natexlab{b}}.

\bibitem[Lumer et~al.(2025)Lumer, Basavaraju, Mason, Burke, and
  Subbiah]{lumer2025graph}
E.~Lumer, P.~H. Basavaraju, M.~Mason, J.~A. Burke, and V.~K. Subbiah.
\newblock Graph {RAG}-tool fusion.
\newblock \emph{arXiv preprint arXiv:2502.07223}, 2025.

\bibitem[Nie et~al.(2026)Nie, Shen, Yu, Yin, Zhang, and Hu]{nie2026skillgraph}
Z.~Nie, R.~Shen, X.~Yu, B.~Yin, J.~Zhang, and X.~Hu.
\newblock {SkillGraph}: Self-evolving multi-agent collaboration with multimodal
  graph topology.
\newblock \emph{arXiv preprint arXiv:2604.17503}, 2026.

\bibitem[Park et~al.(2023)Park, O'Brien, Cai, Morris, Liang, and
  Bernstein]{park2023generative}
J.~S. Park, J.~O'Brien, C.~J. Cai, M.~R. Morris, P.~Liang, and M.~S. Bernstein.
\newblock Generative agents: Interactive simulacra of human behavior.
\newblock In \emph{Proceedings of the 36th annual acm symposium on user
  interface software and technology}, pages 1--22, 2023.

\bibitem[Patil et~al.(2024)Patil, Zhang, Wang, and Gonzalez]{patil2024gorilla}
S.~G. Patil, T.~Zhang, X.~Wang, and J.~E. Gonzalez.
\newblock {Gorilla}: Large language model connected with massive {APIs}.
\newblock \emph{Advances in Neural Information Processing Systems},
  37:\penalty0 126544--126565, 2024.

\bibitem[Patil et~al.(2025)Patil, Mao, Yan, Ji, Suresh, Stoica, and
  Gonzalez]{patil2025berkeley}
S.~G. Patil, H.~Mao, F.~Yan, C.~C.-J. Ji, V.~Suresh, I.~Stoica, and J.~E.
  Gonzalez.
\newblock The {Berkeley} function calling leaderboard ({BFCL}): From tool use
  to agentic evaluation of large language models.
\newblock In \emph{Forty-second International Conference on Machine Learning},
  2025.

\bibitem[Patwardhan et~al.(2025)Patwardhan, Dias, Proehl, Kim, Wang, Watkins,
  Posada~Fishman, Aljubeh, Thacker, Fauconnet, Kim,
  et~al.]{patwardhan2025gdpval}
T.~Patwardhan, R.~Dias, E.~Proehl, G.~Kim, M.~Wang, O.~Watkins,
  S.~Posada~Fishman, M.~Aljubeh, P.~Thacker, L.~Fauconnet, N.~S. Kim, et~al.
\newblock {GDPval}: Evaluating {AI} model performance on real-world
  economically valuable tasks.
\newblock \emph{SuperIntelligence-Robotics-Safety \& Alignment}, 2\penalty0
  (4), 2025.

\bibitem[Prasad et~al.(2024)Prasad, Koller, Hartmann, Clark, Sabharwal, Bansal,
  and Khot]{prasad2024adapt}
A.~Prasad, A.~Koller, M.~Hartmann, P.~Clark, A.~Sabharwal, M.~Bansal, and
  T.~Khot.
\newblock {ADaPT}: As-needed decomposition and planning with language models.
\newblock In \emph{Findings of the Association for Computational Linguistics:
  NAACL 2024}, pages 4226--4252, 2024.

\bibitem[Qin et~al.(2024)Qin, Liang, Ye, Zhu, Yan, Lu, Lin, Cong, Tang, Qian,
  et~al.]{qin2024toolllm}
Y.~Qin, S.~Liang, Y.~Ye, K.~Zhu, L.~Yan, Y.~Lu, Y.~Lin, X.~Cong, X.~Tang,
  B.~Qian, et~al.
\newblock {ToolLLM}: Facilitating large language models to master 16000+
  real-world {APIs}.
\newblock In \emph{International Conference on Learning Representations},
  volume 2024, pages 9695--9717, 2024.

\bibitem[Qu et~al.(2024)Qu, Dai, Wei, Cai, Wang, Yin, Xu, and
  Wen]{qu2024towards}
C.~Qu, S.~Dai, X.~Wei, H.~Cai, S.~Wang, D.~Yin, J.~Xu, and J.-R. Wen.
\newblock Towards completeness-oriented tool retrieval for large language
  models.
\newblock In \emph{Proceedings of the 33rd ACM International Conference on
  Information and Knowledge Management}, pages 1930--1940, 2024.

\bibitem[Rasmussen et~al.(2025)Rasmussen, Paliychuk, Beauvais, Ryan, and
  Chalef]{rasmussen2025zep}
P.~Rasmussen, P.~Paliychuk, T.~Beauvais, J.~Ryan, and D.~Chalef.
\newblock {Zep}: a temporal knowledge graph architecture for agent memory.
\newblock \emph{arXiv preprint arXiv:2501.13956}, 2025.

\bibitem[Schick et~al.(2023)Schick, Dwivedi-Yu, Dess{\`\i}, Raileanu, Lomeli,
  Hambro, Zettlemoyer, Cancedda, and Scialom]{schick2023toolformer}
T.~Schick, J.~Dwivedi-Yu, R.~Dess{\`\i}, R.~Raileanu, M.~Lomeli, E.~Hambro,
  L.~Zettlemoyer, N.~Cancedda, and T.~Scialom.
\newblock {Toolformer}: Language models can teach themselves to use tools.
\newblock \emph{Advances in neural information processing systems},
  36:\penalty0 68539--68551, 2023.

\bibitem[Shen et~al.(2023)Shen, Song, Tan, Li, Lu, and
  Zhuang]{shen2023hugginggpt}
Y.~Shen, K.~Song, X.~Tan, D.~Li, W.~Lu, and Y.~Zhuang.
\newblock {HuggingGPT}: Solving {AI} tasks with {ChatGPT} and its friends in
  {Hugging Face}.
\newblock \emph{Advances in Neural Information Processing Systems},
  36:\penalty0 38154--38180, 2023.

\bibitem[Shen et~al.(2024)Shen, Song, Tan, Zhang, Ren, Yuan, Lu, Li, and
  Zhuang]{shen2024taskbench}
Y.~Shen, K.~Song, X.~Tan, W.~Zhang, K.~Ren, S.~Yuan, W.~Lu, D.~Li, and
  Y.~Zhuang.
\newblock {TaskBench}: Benchmarking large language models for task automation.
\newblock \emph{Advances in Neural Information Processing Systems},
  37:\penalty0 4540--4574, 2024.

\bibitem[Shi et~al.(2026)Shi, Chen, Jiang, Song, Yang, and
  Zhao]{shi2026experiential}
T.~Shi, S.~Chen, B.~Jiang, L.~Song, L.~Yang, and J.~Zhao.
\newblock Experiential reinforcement learning.
\newblock \emph{arXiv preprint arXiv:2602.13949}, 2026.

\bibitem[Shinn et~al.(2023)Shinn, Cassano, Gopinath, Narasimhan, and
  Yao]{shinn2023reflexion}
N.~Shinn, F.~Cassano, A.~Gopinath, K.~Narasimhan, and S.~Yao.
\newblock {Reflexion}: Language agents with verbal reinforcement learning.
\newblock \emph{Advances in neural information processing systems},
  36:\penalty0 8634--8652, 2023.

\bibitem[Shridhar et~al.(2021)Shridhar, Yuan, Cote, Bisk, Trischler, and
  Hausknecht]{shridharalfworld}
M.~Shridhar, X.~Yuan, M.-A. Cote, Y.~Bisk, A.~Trischler, and M.~Hausknecht.
\newblock {ALFWorld}: Aligning text and embodied environments for interactive
  learning.
\newblock In \emph{International Conference on Learning Representations}, 2021.

\bibitem[Sumers et~al.(2023)Sumers, Yao, Narasimhan, and
  Griffiths]{sumers2023cognitive}
T.~Sumers, S.~Yao, K.~R. Narasimhan, and T.~L. Griffiths.
\newblock Cognitive architectures for language agents.
\newblock \emph{Transactions on Machine Learning Research}, 2023.

\bibitem[Sun et~al.(2025)Sun, Liu, Zang, Cao, Dong, Wu, Lin, and
  Wang]{sun2025seagent}
Z.~Sun, Z.~Liu, Y.~Zang, Y.~Cao, X.~Dong, T.~Wu, D.~Lin, and J.~Wang.
\newblock {SE-Agent}: Self-evolving computer use agent with autonomous learning
  from experience.
\newblock \emph{arXiv preprint arXiv:2508.04700}, 2025.

\bibitem[Wang et~al.(2023{\natexlab{a}})Wang, Xie, Jiang, Mandlekar, Xiao, Zhu,
  Fan, and Anandkumar]{wang2023voyager}
G.~Wang, Y.~Xie, Y.~Jiang, A.~Mandlekar, C.~Xiao, Y.~Zhu, L.~Fan, and
  A.~Anandkumar.
\newblock {Voyager}: An open-ended embodied agent with large language models.
\newblock \emph{arXiv preprint arXiv:2305.16291}, 2023{\natexlab{a}}.

\bibitem[Wang et~al.(2023{\natexlab{b}})Wang, Xu, Lan, Hu, Lan, Lee, and
  Lim]{wang2023plan}
L.~Wang, W.~Xu, Y.~Lan, Z.~Hu, Y.~Lan, R.~K.-W. Lee, and E.-P. Lim.
\newblock Plan-and-solve prompting: Improving zero-shot chain-of-thought
  reasoning by large language models.
\newblock In \emph{Proceedings of the 61st annual meeting of the association
  for computational linguistics (volume 1: long papers)}, pages 2609--2634,
  2023{\natexlab{b}}.

\bibitem[Wang et~al.(2025{\natexlab{a}})Wang, Han, Ji, Wang, Baldwin, and
  Li]{wang2025toolgen}
R.~Wang, X.~Han, L.~Ji, S.~Wang, T.~Baldwin, and H.~Li.
\newblock {ToolGen}: Unified tool retrieval and calling via generation.
\newblock In \emph{International Conference on Learning Representations},
  volume 2025, pages 73473--73498, 2025{\natexlab{a}}.

\bibitem[Wang et~al.(2024{\natexlab{a}})Wang, Chen, Yuan, Zhang, Li, Peng, and
  Ji]{wang2024executable}
X.~Wang, Y.~Chen, L.~Yuan, Y.~Zhang, Y.~Li, H.~Peng, and H.~Ji.
\newblock Executable code actions elicit better {LLM} agents.
\newblock \emph{arXiv preprint arXiv:2402.01030}, 2024{\natexlab{a}}.

\bibitem[Wang et~al.(2024{\natexlab{b}})Wang, Fried, and Neubig]{wang2024trove}
Z.~Wang, D.~Fried, and G.~Neubig.
\newblock {TroVE}: Inducing verifiable and efficient toolboxes for solving
  programmatic tasks.
\newblock \emph{arXiv preprint arXiv:2401.12869}, 2024{\natexlab{b}}.

\bibitem[Wang et~al.(2026)Wang, Wu, Zhang, Zhang, Yao, Faisal, Peng, Qin, Nath,
  Lin, et~al.]{wang2026webxskill}
Z.~Wang, Q.~Wu, X.~Zhang, C.~Zhang, W.~Yao, F.~E. Faisal, B.~Peng, S.~Qin,
  S.~Nath, Q.~Lin, et~al.
\newblock {WebXSkill}: Skill learning for autonomous web agents.
\newblock \emph{arXiv preprint arXiv:2604.13318}, 2026.

\bibitem[Wang et~al.(2025{\natexlab{b}})Wang, Mao, Fried, and
  Neubig]{wang2025agent}
Z.~Z. Wang, J.~Mao, D.~Fried, and G.~Neubig.
\newblock Agent workflow memory.
\newblock In \emph{International Conference on Machine Learning}, pages
  63897--63911. PMLR, 2025{\natexlab{b}}.

\bibitem[Willard and Louf(2023)]{willard2023efficient}
B.~T. Willard and R.~Louf.
\newblock Efficient guided generation for large language models.
\newblock \emph{arXiv preprint arXiv:2307.09702}, 2023.

\bibitem[Wu et~al.(2025)Wu, Wang, Mei, Cai, Fu, Yang, Wen, Yang, Shen, Wang,
  et~al.]{wu2025evolver}
R.~Wu, X.~Wang, J.~Mei, P.~Cai, D.~Fu, C.~Yang, L.~Wen, X.~Yang, Y.~Shen,
  Y.~Wang, et~al.
\newblock {EvolveR}: Self-evolving {LLM} agents through an experience-driven
  lifecycle.
\newblock \emph{arXiv preprint arXiv:2510.16079}, 2025.

\bibitem[Wu et~al.(2024)Wu, Shen, Shan, Song, Wang, Zhang, Feng, Cheng, Chen,
  Xiong, et~al.]{wu2024can}
X.~Wu, Y.~Shen, C.~Shan, K.~Song, S.~Wang, B.~Zhang, J.~Feng, H.~Cheng,
  W.~Chen, Y.~Xiong, et~al.
\newblock Can graph learning improve planning in {LLM}-based agents?
\newblock \emph{Advances in Neural Information Processing Systems},
  37:\penalty0 5338--5383, 2024.

\bibitem[Xiao et~al.(2024)Xiao, Ma, Wang, Wu, Zhao, Wang, Huang, and
  Li]{xiao2024flowbench}
R.~Xiao, W.~Ma, K.~Wang, Y.~Wu, J.~Zhao, H.~Wang, F.~Huang, and Y.~Li.
\newblock {FlowBench}: Revisiting and benchmarking workflow-guided planning for
  {LLM}-based agents.
\newblock In \emph{Findings of the Association for Computational Linguistics:
  EMNLP 2024}, pages 10883--10900, 2024.

\bibitem[Xu et~al.(2026)Xu, Liang, Mei, Gao, Tan, and Zhang]{xu2026mem}
W.~Xu, Z.~Liang, K.~Mei, H.~Gao, J.~Tan, and Y.~Zhang.
\newblock {A-Mem}: Agentic memory for {LLM} agents.
\newblock \emph{Advances in Neural Information Processing Systems},
  38:\penalty0 17577--17604, 2026.

\bibitem[Yang et~al.(2018)Yang, Qi, Zhang, Bengio, Cohen, Salakhutdinov, and
  Manning]{yang2018hotpotqa}
Z.~Yang, P.~Qi, S.~Zhang, Y.~Bengio, W.~Cohen, R.~Salakhutdinov, and C.~D.
  Manning.
\newblock {HotpotQA}: A dataset for diverse, explainable multi-hop question
  answering.
\newblock In \emph{Proceedings of the 2018 conference on empirical methods in
  natural language processing}, pages 2369--2380, 2018.

\bibitem[Yao et~al.(2023{\natexlab{a}})Yao, Yu, Zhao, Shafran, Griffiths, Cao,
  and Narasimhan]{yao2023tree}
S.~Yao, D.~Yu, J.~Zhao, I.~Shafran, T.~Griffiths, Y.~Cao, and K.~Narasimhan.
\newblock Tree of thoughts: Deliberate problem solving with large language
  models.
\newblock \emph{Advances in neural information processing systems},
  36:\penalty0 11809--11822, 2023{\natexlab{a}}.

\bibitem[Yao et~al.(2023{\natexlab{b}})Yao, Zhao, Yu, Du, Shafran, Narasimhan,
  and Cao]{yao2023react}
S.~Yao, J.~Zhao, D.~Yu, N.~Du, I.~Shafran, K.~Narasimhan, and Y.~Cao.
\newblock {ReAct}: Synergizing reasoning and acting in language models.
\newblock In \emph{International Conference on Learning Representations
  (ICLR)}, 2023{\natexlab{b}}.

\bibitem[Yao et~al.(2024)Yao, Shinn, Razavi, and Narasimhan]{yao2024tau}
S.~Yao, N.~Shinn, P.~Razavi, and K.~Narasimhan.
\newblock tau-bench: A benchmark for tool-agent-user interaction in real-world
  domains.
\newblock \emph{arXiv preprint arXiv:2406.12045}, 2024.

\bibitem[Zhang et~al.(2025)Zhang, Xiang, Yu, Teng, Chen, Chen, Zhuge, Cheng,
  Hong, Wang, et~al.]{zhang2025aflow}
J.~Zhang, J.~Xiang, Z.~Yu, F.~Teng, X.~Chen, J.~Chen, M.~Zhuge, X.~Cheng,
  S.~Hong, J.~Wang, et~al.
\newblock {AFlow}: Automating agentic workflow generation.
\newblock In \emph{International Conference on Learning Representations},
  volume 2025, pages 34040--34077, 2025.

\bibitem[Zhang et~al.(2023)Zhang, Chen, Li, and Wang]{zhang2023don}
K.~Zhang, H.~Chen, L.~Li, and W.~Wang.
\newblock Don't fine-tune, decode: Syntax error-free tool use via constrained
  decoding.
\newblock \emph{arXiv preprint arXiv:2310.07075}, 2023.

\bibitem[Zhao et~al.(2024)Zhao, Huang, Xu, Lin, Liu, and Huang]{zhao2024expel}
A.~Zhao, D.~Huang, Q.~Xu, M.~Lin, Y.-J. Liu, and G.~Huang.
\newblock {ExpeL}: {LLM} agents are experiential learners.
\newblock In \emph{Proceedings of the AAAI Conference on Artificial
  Intelligence}, volume~38, pages 19632--19642, 2024.

\bibitem[Zheng et~al.(2025)Zheng, Fatemi, Jin, Wang, Gandhi, Song, Gu,
  Srinivasa, Liu, Neubig, et~al.]{zheng2025skillweaver}
B.~Zheng, M.~Y. Fatemi, X.~Jin, Z.~Z. Wang, A.~Gandhi, Y.~Song, Y.~Gu,
  J.~Srinivasa, G.~Liu, G.~Neubig, et~al.
\newblock {SkillWeaver}: Web agents can self-improve by discovering and honing
  skills.
\newblock \emph{arXiv preprint arXiv:2504.07079}, 2025.

\bibitem[Zhong et~al.(2024)Zhong, Guo, Gao, Ye, and Wang]{zhong2024memorybank}
W.~Zhong, L.~Guo, Q.~Gao, H.~Ye, and Y.~Wang.
\newblock {MemoryBank}: Enhancing large language models with long-term memory.
\newblock In \emph{Proceedings of the AAAI conference on artificial
  intelligence}, volume~38, pages 19724--19731, 2024.
\newblock Issue 17.

\bibitem[Zhu et~al.(2025)Zhu, Qiao, Ou, Deng, Lyu, Shen, Liang, Gu, Chen, and
  Zhang]{zhu2025knowagent}
Y.~Zhu, S.~Qiao, Y.~Ou, S.~Deng, S.~Lyu, Y.~Shen, L.~Liang, J.~Gu, H.~Chen, and
  N.~Zhang.
\newblock {KnowAgent}: Knowledge-augmented planning for {LLM}-based agents.
\newblock In \emph{Findings of the Association for Computational Linguistics:
  NAACL 2025}, pages 3709--3732, 2025.

\end{thebibliography}
\end{document}